\documentclass[journal]{IEEEtran}

\usepackage{amsmath,amsfonts}
\usepackage{algorithm}
\usepackage{array}
\usepackage[caption=false,font=normalsize,labelfont=sf,textfont=sf]{subfig}
\usepackage{textcomp}
\usepackage{stfloats}
\usepackage{url}
\usepackage{verbatim}
\usepackage{graphicx}
\usepackage{cite}
\usepackage[T1]{fontenc}
\usepackage[utf8]{inputenc}
\usepackage{mathtools}

\usepackage{amssymb,mathrsfs}
\usepackage{amsthm}
\usepackage{bm}
\usepackage{scalerel}
\usepackage{nicefrac}
\usepackage{microtype} 
\usepackage[shortlabels]{enumitem}
\usepackage{graphicx}
\usepackage{epstopdf}
\DeclareGraphicsExtensions{.eps,.png,.jpg,.pdf}

\usepackage{url}
\usepackage{colortbl}
\usepackage{booktabs}
\usepackage{multirow}
\usepackage{colortbl,xcolor}
\usepackage{soul}
\usepackage{xparse,xstring}
\usepackage{calc}
\usepackage{etoolbox}

\makeatletter
\@ifpackageloaded{natbib}{
	\relax
}{
	\usepackage{cite}
}
\makeatother

\usepackage{array}
\newcolumntype{L}[1]{>{\raggedright\let\newline\\\arraybackslash\hspace{0pt}}m{#1}}
\newcolumntype{C}[1]{>{\centering\let\newline\\\arraybackslash\hspace{0pt}}m{#1}}
\newcolumntype{R}[1]{>{\raggedleft\let\newline\\\arraybackslash\hspace{0pt}}m{#1}}

\usepackage{glossaries}
\makeatletter
\sfcode`\.1006

\let\oldgls\gls
\let\oldglspl\glspl

\newcommand\fussy@ifnextchar[3]{%
	\let\reserved@d=#1%
	\def\reserved@a{#2}%
	\def\reserved@b{#3}%
	\futurelet\@let@token\fussy@ifnch}
\def\fussy@ifnch{%
	\ifx\@let@token\reserved@d
		\let\reserved@c\reserved@a
	\else
		\let\reserved@c\reserved@b
	\fi
	\reserved@c}

\renewcommand{\gls}[1]{%
\oldgls{#1}\fussy@ifnextchar.{\@checkperiod}{\@}}
\renewcommand{\glspl}[1]{%
\oldglspl{#1}\fussy@ifnextchar.{\@checkperiod}{\@}}

\newcommand{\@checkperiod}[1]{%
	\ifnum\sfcode`\.=\spacefactor\else#1\fi
}

\robustify{\gls}
\robustify{\glspl}
\makeatother

\newacronym{wrt}{w.r.t.}{with respect to}
\newacronym{RHS}{R.H.S.}{right-hand side}
\newacronym{LHS}{L.H.S.}{left-hand side}
\newacronym{iid}{i.i.d.}{independent and identically distributed}
\newacronym{SOTA}{SOTA}{state-of-the-art}

\usepackage{float}

\ifx\notloadhyperref\undefined
	\ifx\loadbibentry\undefined
		\usepackage[hidelinks,hypertexnames=false]{hyperref}
	\else
		\usepackage{bibentry}
		\makeatletter\let\saved@bibitem\@bibitem\makeatother
		\usepackage[hidelinks,hypertexnames=false]{hyperref}
		\makeatletter\let\@bibitem\saved@bibitem\makeatother
	\fi
\else
	\ifx\loadbibentry\undefined
		\relax
	\else
		\usepackage{bibentry}
	\fi
\fi

\usepackage[capitalize]{cleveref}
\makeatletter
\def\cref@getref#1#2{%
  \expandafter\let\expandafter#2\csname r@#1@cref\endcsname%
  \expandafter\expandafter\expandafter\def%
    \expandafter\expandafter\expandafter#2%
    \expandafter\expandafter\expandafter{%
      \expandafter\@firstoffive#2}}
\def\cpageref@getref#1#2{%
  \expandafter\let\expandafter#2\csname r@#1@cref\endcsname%
  \expandafter\expandafter\expandafter\def%
    \expandafter\expandafter\expandafter#2%
    \expandafter\expandafter\expandafter{%
      \expandafter\@secondoffive#2}}
      
\AtBeginDocument{%
   \def\label@noarg#1{%
    \cref@old@label{#1}%
    \@bsphack%
    \edef\@tempa{{page}{\the\c@page}}%
    \setcounter{page}{1}%
    \edef\@tempb{\thepage}%
    \expandafter\setcounter\@tempa%
    \cref@constructprefix{page}{\cref@result}%
    \protected@write\@auxout{}%
      {\string\newlabel{#1@cref}{{\cref@currentlabel}%
      {[\@tempb][\arabic{page}][\cref@result]\thepage}{}{}{}}}
    \@esphack}%
  \def\label@optarg[#1]#2{%
    \cref@old@label{#2}%
    \@bsphack%
    \edef\@tempa{{page}{\the\c@page}}%
    \setcounter{page}{1}%
    \edef\@tempb{\thepage}%
    \expandafter\setcounter\@tempa%
    \cref@constructprefix{page}{\cref@result}%
    \protected@edef\cref@currentlabel{%
      \expandafter\cref@override@label@type%
        \cref@currentlabel\@nil{#1}}%
    \protected@write\@auxout{}%
      {\string\newlabel{#2@cref}{{\cref@currentlabel}%
      {[\@tempb][\arabic{page}][\cref@result]\thepage}{}{}{}}}
    \@esphack}%
    }  
\makeatother

\crefname{equation}{}{}
\Crefname{equation}{}{}
\crefname{claim}{claim}{claims}
\crefname{step}{step}{steps}
\crefname{line}{line}{lines}
\crefname{condition}{condition}{conditions}
\crefname{dmath}{}{}
\crefname{dseries}{}{}
\crefname{dgroup}{}{}
\crefname{page}{page}{pages}

\crefname{Problem}{Problem}{Problems}
\crefformat{Problem}{Problem~#2#1#3}
\crefrangeformat{Problem}{Problems~#3#1#4 to~#5#2#6}

\crefname{Theorem}{Theorem}{Theorems}
\crefname{Corollary}{Corollary}{Corollaries}
\crefname{Proposition}{Proposition}{Propositions}
\crefname{Lemma}{Lemma}{Lemmas}
\crefname{Definition}{Definition}{Definitions}
\crefname{Example}{Example}{Examples}
\crefname{Assumption}{Assumption}{Assumptions}
\crefname{Remark}{Remark}{Remarks}
\crefname{Rem}{Remark}{Remarks}
\crefname{remarks}{Remarks}{Remarks}
\crefname{Appendix}{Appendix}{Appendices}
\crefname{Supplement}{Supplement}{Supplements}
\crefname{Exercise}{Exercise}{Exercises}
\crefname{Theorem_A}{Theorem}{Theorems}
\crefname{Corollary_A}{Corollary}{Corollaries}
\crefname{Proposition_A}{Proposition}{Propositions}
\crefname{Lemma_A}{Lemma}{Lemmas}
\crefname{Definition_A}{Definition}{Definitions}

\usepackage{crossreftools}
\ifx\notloadhyperref\undefined
\else
	\relax
\fi

\usepackage{algorithm}
\usepackage{algpseudocode}

\ifx\loadbreqn\undefined
	\relax
\else
	\usepackage{breqn}
\fi

\makeatletter
\def\cleartheorem#1{%
    \expandafter\let\csname#1\endcsname\relax
    \expandafter\let\csname c@#1\endcsname\relax
}
\def\clearthms#1{ \@for\tname:=#1\do{\cleartheorem\tname} }
\makeatother

\ifx\renewtheorem\undefined
	\ifx\useTheoremCounter\undefined
		\newtheorem{Theorem}{Theorem}
		\newtheorem{Corollary}{Corollary}
		\newtheorem{Proposition}{Proposition}
		
	\else

	\fi

\fi

\theoremstyle{remark}

\theoremstyle{plain}

\newcommand{\qednew}{\nobreak \ifvmode \relax \else
		\ifdim\lastskip<1.5em \hskip-\lastskip
			\hskip1.5em plus0em minus0.5em \fi \nobreak
		\vrule height0.75em width0.5em depth0.25em\fi}



\newcommand{\ml}[1]{\begin{multlined}[t]#1\end{multlined}}

\NewDocumentCommand{\movedownsub}{e{^_}}{%
	\IfNoValueTF{#1}{%
		\IfNoValueF{#2}{^{}}
	}{%
		^{#1}
	}%
	\IfNoValueF{#2}{_{#2}}
}

\let\latexchi\chi
\RenewDocumentCommand{\chi}{}{\latexchi\movedownsub}

\newcommand{\Real}{\mathbb{R}}

\newcommand{\calF}{\mathcal{F}}

\newcommand{\calM}{\mathcal{M}}
\newcommand{\calN}{\mathcal{N}}

\newcommand{\bA}{\mathbf{A}}

\newcommand{\bE}{\mathbf{E}}
\newcommand{\boldf}{\mathbf{f}}
\newcommand{\bF}{\mathbf{F}}
\newcommand{\bg}{\mathbf{g}}
\newcommand{\bG}{\mathbf{G}}

\newcommand{\bI}{\mathbf{I}}

\newcommand{\bK}{\mathbf{K}}

\newcommand{\bp}{\mathbf{p}}
\newcommand{\bP}{\mathbf{P}}

\newcommand{\bQ}{\mathbf{Q}}
\newcommand{\br}{\mathbf{r}}
\newcommand{\bR}{\mathbf{R}}
\newcommand{\bs}{\mathbf{s}}

\newcommand{\bt}{\mathbf{t}}

\newcommand{\bu}{\mathbf{u}}

\newcommand{\bv}{\mathbf{v}}
\newcommand{\bV}{\mathbf{V}}

\newcommand{\bW}{\mathbf{W}}
\newcommand{\bx}{\mathbf{x}}
\newcommand{\bX}{\mathbf{X}}
\newcommand{\by}{\mathbf{y}}
\newcommand{\bY}{\mathbf{Y}}
\newcommand{\bz}{\mathbf{z}}
\newcommand{\bZ}{\mathbf{Z}}

\newcommand{\mby}{\bm{y}}

\newcommand{\bbR}{\mathbb{R}}

\DeclareSymbolFont{bsfletters}{OT1}{cmss}{bx}{n}
\DeclareSymbolFont{ssfletters}{OT1}{cmss}{m}{n}
\DeclareMathSymbol{\bsfGamma}{0}{bsfletters}{'000}
\DeclareMathSymbol{\ssfGamma}{0}{ssfletters}{'000}
\DeclareMathSymbol{\bsfDelta}{0}{bsfletters}{'001}
\DeclareMathSymbol{\ssfDelta}{0}{ssfletters}{'001}
\DeclareMathSymbol{\bsfTheta}{0}{bsfletters}{'002}
\DeclareMathSymbol{\ssfTheta}{0}{ssfletters}{'002}
\DeclareMathSymbol{\bsfLambda}{0}{bsfletters}{'003}
\DeclareMathSymbol{\ssfLambda}{0}{ssfletters}{'003}
\DeclareMathSymbol{\bsfXi}{0}{bsfletters}{'004}
\DeclareMathSymbol{\ssfXi}{0}{ssfletters}{'004}
\DeclareMathSymbol{\bsfPi}{0}{bsfletters}{'005}
\DeclareMathSymbol{\ssfPi}{0}{ssfletters}{'005}
\DeclareMathSymbol{\bsfSigma}{0}{bsfletters}{'006}
\DeclareMathSymbol{\ssfSigma}{0}{ssfletters}{'006}
\DeclareMathSymbol{\bsfUpsilon}{0}{bsfletters}{'007}
\DeclareMathSymbol{\ssfUpsilon}{0}{ssfletters}{'007}
\DeclareMathSymbol{\bsfPhi}{0}{bsfletters}{'010}
\DeclareMathSymbol{\ssfPhi}{0}{ssfletters}{'010}
\DeclareMathSymbol{\bsfPsi}{0}{bsfletters}{'011}
\DeclareMathSymbol{\ssfPsi}{0}{ssfletters}{'011}
\DeclareMathSymbol{\bsfOmega}{0}{bsfletters}{'012}
\DeclareMathSymbol{\ssfOmega}{0}{ssfletters}{'012}

\makeatletter
\newcommand*\rel@kern[1]{\kern#1\dimexpr\macc@kerna}
\newcommand*\widebar[1]{%
  \begingroup
  \def\mathaccent##1##2{%
    \rel@kern{0.8}%
    \overline{\rel@kern{-0.8}\macc@nucleus\rel@kern{0.2}}%
    \rel@kern{-0.2}%
  }%
  \macc@depth\@ne
  \let\math@bgroup\@empty \let\math@egroup\macc@set@skewchar
  \mathsurround\z@ \frozen@everymath{\mathgroup\macc@group\relax}%
  \macc@set@skewchar\relax
  \let\mathaccentV\macc@nested@a
  \macc@nested@a\relax111{#1}%
  \endgroup
}
\makeatother

\DeclareMathOperator{\var}{var}

\DeclareMathOperator{\cov}{cov}

\DeclareMathOperator*{\concat}{\scalerel*{\parallel}{\sum}}

\newcommand{\ifbcdot}[1]{\ifblank{#1}{\cdot}{#1}}

\DeclarePairedDelimiterX\abs[1]{\lvert}{\rvert}{\ifbcdot{#1}}
\DeclarePairedDelimiterX\parens[1]{(}{)}{\ifbcdot{#1}}
\DeclarePairedDelimiterX\brk[1]{[}{]}{\ifbcdot{#1}}
\DeclarePairedDelimiterX\braces[1]{\{}{\}}{\ifbcdot{#1}}
\DeclarePairedDelimiterX\angles[1]{\langle}{\rangle}{\ifblank{#1}{\cdot,\cdot}{#1}}
\DeclarePairedDelimiterX\ip[2]{\langle}{\rangle}{\ifbcdot{#1},\ifbcdot{#2}}
\DeclarePairedDelimiterX\norm[1]{\lVert}{\rVert}{\ifbcdot{#1}}
\DeclarePairedDelimiterX\ceil[1]{\lceil}{\rceil}{\ifbcdot{#1}}
\DeclarePairedDelimiterX\floor[1]{\lfloor}{\rfloor}{\ifbcdot{#1}}

\DeclareFontFamily{U}{matha}{\hyphenchar\font45}
\DeclareFontShape{U}{matha}{m}{n}{
      <5> <6> <7> <8> <9> <10> gen * matha
      <10.95> matha10 <12> <14.4> <17.28> <20.74> <24.88> matha12
      }{}
\DeclareSymbolFont{matha}{U}{matha}{m}{n}
\DeclareFontSubstitution{U}{matha}{m}{n}

\DeclareFontFamily{U}{mathx}{\hyphenchar\font45}
\DeclareFontShape{U}{mathx}{m}{n}{
      <5> <6> <7> <8> <9> <10>
      <10.95> <12> <14.4> <17.28> <20.74> <24.88>
      mathx10
      }{}
\DeclareSymbolFont{mathx}{U}{mathx}{m}{n}
\DeclareFontSubstitution{U}{mathx}{m}{n}

\DeclareMathDelimiter{\vvvert}{0}{matha}{"7E}{mathx}{"17}
\DeclarePairedDelimiterX\vertiii[1]{\vvvert}{\vvvert}{\ifbcdot{#1}}

\DeclarePairedDelimiterXPP\trace[1]{\operatorname{Tr}}{(}{)}{}{\ifbcdot{#1}} 
\DeclarePairedDelimiterXPP\col[1]{\operatorname{col}}{\{}{\}}{}{\ifbcdot{#1}} 
\DeclarePairedDelimiterXPP\row[1]{\operatorname{row}}{\{}{\}}{}{\ifbcdot{#1}} 
\DeclarePairedDelimiterXPP\erf[1]{\operatorname{erf}}{(}{)}{}{\ifbcdot{#1}}
\DeclarePairedDelimiterXPP\erfc[1]{\operatorname{erfc}}{(}{)}{}{\ifbcdot{#1}}
\DeclarePairedDelimiterXPP\KLD[2]{D}{(}{)}{}{\ifbcdot{#1}\, \delimsize\|\, \ifbcdot{#2}} 
\DeclarePairedDelimiterXPP\op[2]{\operatorname{#1}}{(}{)}{}{#2} 

\newcommand{\T}{^{\mkern-1.5mu\mathop\intercal}}

\DeclarePairedDelimiterXPP\indicate[1]{{\bf 1}}{\{}{\}}{}{\ifbcdot{#1}}

\NewDocumentCommand\ofrac{s m}{%
	\IfBooleanTF#1%
	{\dfrac{1}{#2}}%
	{\frac{1}{#2}}%
}
\NewDocumentCommand\ddfrac{s m m}{%
	\IfBooleanTF#1%
	{\dfrac{\mathrm{d} {#2}}{\mathrm{d} {#3}}}%
	{\frac{\mathrm{d} {#2}}{\mathrm{d} {#3}}}%
}
\NewDocumentCommand\ppfrac{s m m}{%
	\IfBooleanTF#1%
	{\dfrac{\partial {#2}}{\partial {#3}}}%
	{\frac{\partial {#2}}{\partial {#3}}}%
}

\providecommand\given{}

\DeclarePairedDelimiterX\Set[2]\{\}{%
\renewcommand\given{\SetSymbol[\delimsize]{#1}}
#2
}
\DeclarePairedDelimiterX\Setc[1]\{\}{%
\renewcommand\given{\SetSymbol{:}}
#1
}

\NewDocumentCommand\set{s o m}{%
	\IfBooleanTF#1%
	{\IfValueTF{#2}{\Set*{#2}{#3}}{\Setc*{#3}}}%
	{\IfValueTF{#2}{\Set{#2}{#3}}{\Setc{#3}}}%
}

\NewDocumentCommand{\evalat}{ s O{\big} m e{_^} }{%
\IfBooleanTF{#1}%
{\left. #3 \right|}{#3#2|}%
\IfValueT{#4}{_{#4}}%
\IfValueT{#5}{^{#5}}%
}

\providecommand\given{}
\DeclarePairedDelimiterXPP\cprob[1]{}(){}{
\renewcommand\given{\nonscript\,\delimsize\vert\allowbreak\nonscript\,\mathopen{}}%
#1%
}
\DeclarePairedDelimiterXPP\cexp[1]{}[]{}{
\renewcommand\given{\nonscript\,\delimsize\vert\allowbreak\nonscript\,\mathopen{}}%
#1%
}

\DeclareDocumentCommand \P { s e{_^} d() g } {%
	\mathbb{P}%
	\IfBooleanTF{#1}%
		{
			\IfValueT{#2}{_{#2}}%
			\IfValueT{#3}{^{#3}}%
			\IfValueTF{#5}{\cprob{#4 \given #5}}{\IfValueT{#4}{\cprob{#4}}}%
		}%
		{
			\IfValueT{#2}{_{#2}}%
			\IfValueT{#3}{^{#3}}%
			\IfValueTF{#5}{\cprob*{#4 \given #5}}{\IfValueT{#4}{\cprob*{#4}}}%
		}%
}

\DeclareDocumentCommand \E { s e{_^} o g } {%
	\mathbb{E}%
	\IfBooleanTF{#1}%
		{
			\IfValueT{#2}{_{#2}}%
			\IfValueT{#3}{^{#3}}%
			\IfValueTF{#5}{\cexp{#4 \given #5}}{\IfValueT{#4}{\cexp{#4}}}%
		}%
		{
			\IfValueT{#2}{_{#2}}%
			\IfValueT{#3}{^{#3}}%
			\IfValueTF{#5}{\cexp*{#4 \given #5}}{\IfValueT{#4}{\cexp*{#4}}}%
		}%
}

\DeclareDocumentCommand \Var { s e{_^} d() g } {%
	\var%
	\IfBooleanTF{#1}%
		{
			\IfValueT{#2}{_{#2}}%
			\IfValueT{#3}{^{#3}}%
			\IfValueTF{#5}{\cprob{#4 \given #5}}{\IfValueT{#4}{\cprob{#4}}}%
		}%
		{
			\IfValueT{#2}{_{#2}}%
			\IfValueT{#3}{^{#3}}%
			\IfValueTF{#5}{\cprob*{#4 \given #5}}{\IfValueT{#4}{\cprob*{#4}}}%
		}%
}

\DeclareDocumentCommand \Cov { s e{_^} d() g } {%
	\cov%
	\IfBooleanTF{#1}%
		{
			\IfValueT{#2}{_{#2}}%
			\IfValueT{#3}{^{#3}}%
			\IfValueTF{#5}{\cprob{#4 \given #5}}{\IfValueT{#4}{\cprob{#4}}}%
		}%
		{
			\IfValueT{#2}{_{#2}}%
			\IfValueT{#3}{^{#3}}%
			\IfValueTF{#5}{\cprob*{#4 \given #5}}{\IfValueT{#4}{\cprob*{#4}}}%
		}%
}

\ExplSyntaxOn
\NewDocumentCommand \dist {m o o} {%
\mathrm{#1}\left(%
	\IfValueT{#3}{%
		\tl_if_blank:nTF{ #3 }{\cdot\, \middle|\, }{#3\, \middle|\, }%
	}
	\IfValueT{#2}{#2}%
\right)%
}
\ExplSyntaxOff

\NewDocumentCommand {\cbrace} {t+ D[]{black} D(){\widthof{#5}} m m } {%
	\begingroup%
		\color{#2}
		\IfBooleanTF{#1}{%
			\overbrace{#4}^%
		}{
			\underbrace{#4}_%
		}%
		{\parbox[c]{#3}{\centering\footnotesize{#5}}}%
	\endgroup%
}

\let\oldforall\forall
\renewcommand{\forall}{\oldforall \, }

\let\oldexist\exists
\renewcommand{\exists}{\oldexist \, }

\makeatletter

\newcommand{\rankcolor}[2]{%
	\expandafter\renewcommand\csname #1\endcsname[1]{%
		\ifblank{##1}{%
			{\color{#2} \textbf{#2}}%
		}{%
			\ifmmode
				\textcolor{#2}{\bm{##1}}%
			\else%
				{\color{#2} \textbf{##1}}%
			\fi	
		}%
	}
}

\providecommand{\first}{}
\providecommand{\second}{}

\rankcolor{first}{red}
\rankcolor{second}{blue}
\rankcolor{third}{cyan}
\makeatother

\graphicspath{{./Figures/}{./figures/}}
\DeclareDocumentCommand{\includeCroppedPdf}{ o O{./Figures/} m }{
	\IfFileExists{#2#3-crop.pdf}{}{%
		\immediate\write18{pdfcrop #2#3.pdf #2#3-crop.pdf}}%
	\includegraphics[#1]{#2#3-crop.pdf}
}

\makeatletter
\newcommand*{\addFileDependency}[1]{
  \typeout{(#1)}
  \@addtofilelist{#1}
  \IfFileExists{#1}{}{\typeout{No file #1.}}
}
\makeatother

\definecolor{gray90}{gray}{0.9}
\def\colorlist{red,blue,brown,cyan,darkgray,gray,lightgray,green,lime,magenta,olive,orange,pink,purple,teal,violet,white,yellow}

\makeatletter
\def\startcomment{[}
\ifx\nohighlights\undefined
	\newcommand{\createcolor}[1]{%
			\expandafter\newcommand\csname #1\endcsname[1]{{\color{#1} ##1}}%
	}
	\newcommand{\msout}[1]{\text{\color{green} \st{\ensuremath{#1}}}}
	\newcommand{\del}[1]{{\color{green}\ifmmode \msout{#1}\else\st{#1}\fi}}
\else
	\newcommand{\createcolor}[1]{%
			\expandafter\newcommand\csname #1\endcsname[1]{%
				\noexpandarg%
				\StrChar{##1}{1}[\firstletter]%
				\if\firstletter\startcomment%
					\relax
				\else%
					##1
				\fi
			}%
	}
	\newcommand{\msout}[1]{}
	\newcommand{\del}[1]{}
\fi

\def\@tempa#1,{%
    \ifx\relax#1\relax\else
        \createcolor{#1}%
        \expandafter\@tempa
    \fi
}
\expandafter\@tempa\colorlist,\relax,
\makeatother

\newcommand{\hhide}[1]{}

\ifx\diagnoselabel\undefined
	\relax
\else
	\makeatletter
	\def\@testdef #1#2#3{%
		\def\reserved@a{#3}\expandafter \ifx \csname #1@#2\endcsname
			\reserved@a  \else
			\typeout{^^Jlabel #2 changed:^^J%
				\meaning\reserved@a^^J%
				\expandafter\meaning\csname #1@#2\endcsname^^J}%
			\@tempswatrue \fi}
	\makeatother
\fi

\usepackage{makecell}

\renewcommand{\first}[1]{\ifblank{#1}{\textbf{bold}}{\textbf{#1}}}
\renewcommand{\second}[1]{\ifblank{#1}{\underline{underlined}}{\underline{#1}}}

\begin{document}

\title{PRFusion: Toward Effective and Robust Multi-Modal Place Recognition with Image and Point Cloud Fusion}

\author{
Sijie~Wang*,
Qiyu~Kang*, 
Rui~She$^{\dag}$\thanks{$\dag$  Corresponding author: Rui She.},
Kai~Zhao,
Yang~Song,
and
Wee~Peng~Tay,~\IEEEmembership{Senior~Member,~IEEE}
\thanks{* These authors contribute equally.}
\thanks{This research is supported by the Singapore Ministry of Education Academic Research Fund Tier 2 grant MOE-T2EP20220-0002. The computational work for this article was partially performed on resources of the National Supercomputing Centre, Singapore (https://www.nscc.sg).}
\thanks{Sijie Wang, Qiyu Kang, Rui She, Kai Zhao, and Wee
Peng Tay are with the School of Electrical and Electronic Engineering,
Nanyang Technological University, Singapore (e-mail: wang1679@e.ntu.edu.sg; qiyu.kang@ntu.edu.sg; rui.she@ntu.edu.sg; kai.zhao@ntu.edu.sg; wptay@ntu.edu.sg). 
Yang Song is with C3.AI (e-mail: yang.song@connect.polyu.hk).}
}




\maketitle

\begin{abstract}
Place recognition plays a crucial role in the fields of robotics and computer vision, finding applications in areas such as autonomous driving, mapping, and localization. Place recognition identifies a place using query sensor data and a known database. One of the main challenges is to develop a model that can deliver accurate results while being robust to environmental variations. We propose two multi-modal place recognition models, namely PRFusion and PRFusion++. PRFusion utilizes global fusion with manifold metric attention, enabling effective interaction between features without requiring camera-LiDAR extrinsic calibrations. In contrast, PRFusion++ assumes the availability of extrinsic calibrations and leverages pixel-point correspondences to enhance feature learning on local windows. Additionally, both models incorporate neural diffusion layers, which enable reliable operation even in challenging environments. We verify the state-of-the-art performance of both models on three large-scale benchmarks. Notably, they outperform existing models by a substantial margin of +3.0 AR@1 on the demanding Boreas dataset. Furthermore, we conduct ablation studies to validate the effectiveness of our proposed methods. The codes are available at: \url{https://github.com/sijieaaa/PRFusion}
\end{abstract}


\section{Introduction}\label{sec:introduction}
\begin{figure}[!tb]
\begin{center}
\includegraphics[width=0.47\textwidth]{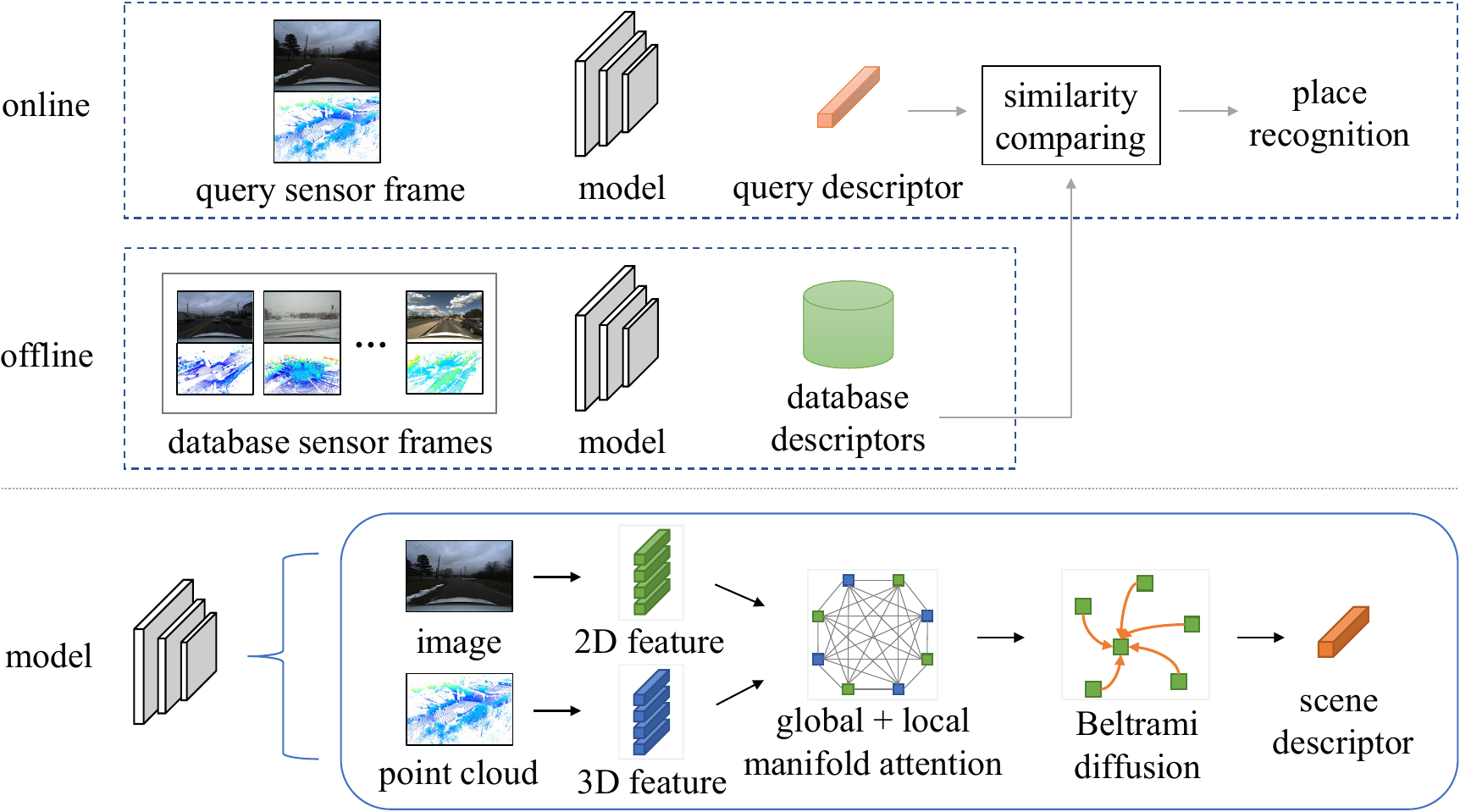}
\end{center}
\caption{Our multi-modal PR pipeline. The query place is recognized by computing a scene descriptor based on both 2D and 3D features using our proposed PR model and then comparing it with the database descriptors. Our proposed model consists of global local feature fusion and neural Beltrami diffusion.}
\label{fig:pipeline}
\end{figure}

\IEEEPARstart{P}{lace}
recognition (PR), a crucial task in computer vision, tries to answer the simple question, ``\textit{Where is it?}''. PR aims to identify a previously visited location stored in a database by comparing query sensor frames (such as images and point clouds) and computing their similarity, as depicted in the pipeline shown in \cref{fig:pipeline}. PR plays a key role in many applications, such as navigation \cite{wang2018navigation}, autonomous driving \cite{doan2019autonomousdriving}, and augmented reality \cite{sarlin2022lamar}. In practical scenarios, it is essential for PR models to operate in real-time during inference while maintaining low memory and storage requirements. To achieve this goal, PR models need to compress the scene frame into a highly summarized descriptor that has a much smaller dimension. For example, a 256$\times$256$\times$3 image is to be summarized as a 512-dimensional feature vector. 

Traditionally, hand-crafted feature techniques are used to aggregate the whole image, e.g., the Bag of Words (BoW)~\cite{galvez2012bagofwords} and Vector of Locally Aggregated Descriptor (VLAD)~\cite{jegou2011vlad}. These solutions depend heavily on prior human knowledge and may not perform well in challenging scenes. Recently, convolutional neural networks (CNNs) have demonstrated a strong capacity in computer vision with learnable convolution layers. NetVLAD \cite{arandjelovic2016netvlad} is the first to leverage learnable layers in PR. It uses CNN to achieve feature extraction, and a learnable VLAD layer to aggregate the image feature map into a single vector representation. Subsequently, a large number of learning-based PR models have been proposed~\cite{radenovic2018gempooling,berton2022cosplace,ali2022convap,ali2023mixvpr}. Despite the effectiveness of image PR models in normal environmental conditions, they usually suffer from poor performance in challenging environments, since cameras are sensitive to environmental illumination changes. By contrast, LiDARs construct a sparse point cloud representation of the surroundings by actively casting laser beams and are more robust against illumination changes. With the popularity of LiDARs in the field of autonomous driving, there have been an increasing number of models leveraging point clouds to tackle the PR problem~\cite{komorowski2021minkloc3d,2021soenet,minkloc3dv2}. Recently, multi-modal fusion using information from both cameras and LiDARs has been proposed. MinkLoc++ and AdaFusion \cite{komorowski2021minkloc++,lai2022adafusion} use separate branches for 2D image and 3D point cloud feature extraction, the outputs of which comprise the final scene descriptor. 

The above models MinkLoc++\cite{komorowski2021minkloc++} and AdaFusion\cite{lai2022adafusion} focus on feature fusion after the aggregation stage, where only the pooled feature descriptors are fed into the fusion module. In addition, they do not consider local interaction, and pixel-point correspondences are not utilized. Furthermore, these approaches do not include any mechanisms to deal robustly with different noisy conditions.

To achieve more effective fusion in this work, we consider global fusion on a more detailed scale, where dense features before the aggregation layer are interrelated. 
We incorporate a learnable manifold metric that supports more effective utilization of multi-modal attention. A manifold is a type of topological space that locally resembles the Euclidean space near each point. Within each local tangent space, the manifold metric supports the definition of the dot product, and the attention \cite{vaswani2017transformer} is a special case of the dot product with a fixed identity manifold metric. However, in the case of multi-modal fusion, vanilla attention based on a static metric may encounter difficulties in accommodating the growing diversity of fusion features. To address this challenge, we propose to incorporate a learnable manifold metric, allowing for more effective utilization of multi-modal attention. 

Moreover, in a further enhanced model, we leverage the pixel-point correspondences to achieve fine-grained feature interaction, where the camera-LiDAR extrinsic calibration that establishes the correspondences is used. Finally, to support the deployment in challenging scenarios, we additionally equip the model with a neural diffusion module to achieve robust feature extraction. The pipeline of our model is shown in \cref{fig:pipeline}. 
Our contributions are summarized as follows:
\begin{itemize}
\item We propose PRFusion, which leverages the Global Fusion Module (GFM) to enable detailed global multi-modal information interaction and integrates metric attention to achieve effective feature fusion. Additionally, PRFusion is equipped with the Neural Diffusion Module (NDM) to enhance feature robustness.
\item We further introduce PRFusion++, which achieves pixel-point level multi-modal feature interaction by incorporating the Local Fusion Module (LFM) with camera-LiDAR extrinsic calibration information.
\item We test our models on large-scale benchmarks and verify that both models achieve state-of-the-art (SOTA) performance. We conduct ablation studies to demonstrate the effectiveness of the modules. 
\end{itemize}

The rest of this paper is organized as follows. 
In \cref{sec:related_work}, we provide a literature review of related work and introduce necessary preliminaries.
In \cref{sec:prfuion}, we present the PRFusion model, which leverages manifold metric attention to achieve global multi-modal feature interaction. It also leverages robust neural diffusion modules.  
Subsequently, in \cref{sec:prfusion++}, we introduce an enhanced model, PRFusion++, which is applicable when camera-LiDAR extrinsic parameters are known, and which incorporates fine-grained local feature fusion. 
In \cref{sec:experiments}, we present a series of experiments that validate the effectiveness of our proposed models.
We conclude and discuss the limitations of our work in \cref{sec:conclusion}.

\section{Related Work and Preliminary}\label{sec:related_work}
In this section, we give an overview of the related PR models in the literature. We then briefly review basic concepts from manifold theory and neural differential diffusion. Finally, we present our problem formulation.

\subsection{PR Models}
\subsubsection{Image PR}

Images produced by cameras are the most commonly used data formats in the problem of PR \cite{she2023image}. Traditionally, hand-crafted models are used to extract scene descriptors from images, such BoW\cite{galvez2012bagofwords} and VLAD\cite{jegou2011vlad}. However, these methods heavily rely on prior information, which may not perform well in challenging environments. To address the limitations of traditional methods, advances in deep learning have inspired learning-based PR models. NetVLAD~\cite{arandjelovic2016netvlad} pioneers the combination of the traditional VLAD descriptor with a CNN to construct a learnable aggregation layer. Its success has paved the way for many image-based PR models. 

Building on the aggregation concept from NetVLAD, there are a series of works \cite{radenovic2018gempooling,ali2022convap,ali2023mixvpr} extending the idea further. For example, GeM~\cite{radenovic2018gempooling} combines max and average pooling operations into a generalized pooling layer with learnable parameters. Based on GeM, ConvAP\cite{ali2022convap} and MixVPR\cite{ali2023mixvpr} further enhance the aggregation strategies by considering holistic feature maps through convolutional layers or multi-layer perceptrons (MLPs). 

In contrast to these aggregation-focused solutions, other works design holistic feature extraction pipelines \cite{hausler2021patchnetvlad, cai2022patchnetvlad+, wang2022transvpr,berton2022cosplace,zhu2023r2former}. For example, PatchNetVLAD \cite{hausler2021patchnetvlad,cai2022patchnetvlad+} advances the strengths of both local and global descriptor methods by deriving patch-level features from NetVLAD residuals. R2Former \cite{zhu2023r2former} combines retrieval and re-ranking procedures to construct more seamless scene descriptors.

However, one of the evident drawbacks of utilizing images is their susceptibility to environmental changes, such as fluctuating illumination and varying weather conditions. 
In this work, we fuse 3D features from a LiDAR sensor with image features to achieve more robust PR performance.


\subsubsection{Point Cloud PR}

In contrast to images, point clouds produced by LiDARs demonstrate greater robustness against environmental perturbations \cite{Wang2023HypLiLoc,li2023sgloc}. Since LiDARs actively emit laser beams, they are less susceptible to variations in illumination.

PointNetVLAD~\cite{uy2018pointnetvlad} marks a significant advancement by utilizing point clouds rather than images for PR. It extracts point cloud features using PointNet \cite{qi2017pointnet,qi2017pointnet++}, which are then processed by a NetVLAD layer to generate the final global descriptor of the scene. This approach has inspired a series of 3D-based models aimed at enhancing point cloud PR.

Following the success of PointNetVLAD, several works have leveraged PointNet-series networks for point-based feature extraction \cite{zhang2019pcan,du2020dh3d,2019lpdnet,2022epcnet}. These methods build on the foundational PointNet architecture \cite{qi2017pointnet,qi2017pointnet++}, optimizing and expanding its capabilities to handle the inherent 3D properties of point clouds.

In addition to point-based methods, voxel-based (cube-based) feature extraction has gained traction \cite{komorowski2021minkloc3d,minkloc3dv2,2021soenet,chen2023ptcnet}. For instance, MinkLoc3D \cite{komorowski2021minkloc3d} pioneers the use of sparse 3D convolutional layers to extract voxelized point cloud features. Building on this, PTC-Net \cite{chen2023ptcnet} extends voxel features with point-wise transformers, providing enhanced scene representations by integrating local and global information more effectively.

Another direction involves projecting the 3D point cloud into 2D formats, employing spherical projection \cite{yin2021psematch,zywanowski2021minkloc3dsi,ma2022overlaptransformer} and birds'-eye-view (BEV) methods \cite{luo2023bevplace}. These projection techniques offer alternative ways to interpret 3D data by leveraging the rich information available in 2D representations, facilitating efficient feature extraction and matching.

However, relying solely on LiDAR sensors overlooks certain perceptual information about street environments, such as the colors of scenes. Therefore, in this work, we incorporate image features to achieve a more comprehensive scene understanding.



\subsubsection{Multi-Modal PR}

Recent research has demonstrated that incorporating multiple modalities can improve performance compared to using a single one. This has led to a growing interest in multi-modal learning using both images and point clouds for PR. The process of multi-modal PR typically employs a two-branch design, where images and point clouds are processed through their respective backbones for basic feature extraction. Then the extracted multi-modal features are fused with fusion layers.

Some multi-modal PR works \cite{lu2020picnet,oertel2020cues,komorowski2021minkloc++,lai2022adafusion} focus on fusing multi-modal features at the global level, often neglecting fine-grained local information. For instance, MinkLoc++ and AdaFusion perform multi-modal feature fusion only on globally-pooled descriptors.

To address this limitation, there are other solutions \cite{zhou2023lcpr,garcia2024umf} that attempt to incorporate fine-grained level information by fully integrating the entire feature maps extracted by the two branches. Specifically, LCPR \cite{zhou2023lcpr} leverages self-attention to fuse image features and spherical-projected point cloud features, while UMF \cite{garcia2024umf} employs both self- and cross-attention mechanisms to capture patterns within both local and global contexts.

Despite these advancements, existing multi-modal PR models still rely on simplistic fusion techniques, such as concatenation \cite{komorowski2021minkloc++} and basic attention mechanisms \cite{lai2022adafusion,zhou2023lcpr,garcia2024umf}. Moreover, they do not exploit camera-LiDAR extrinsic parameters to facilitate effective local fusion. There remains significant potential to enhance multi-modal feature interaction and improve overall performance.

\subsection{Manifold and Metric}
This paper utilizes the mathematical concept of manifolds to provide a theoretical foundation for our proposed approach. 
A manifold $\calM$, of dimension $d$, can be regarded as a generalization of $d$-dimensional Euclidean space, denoted as $\mathbb{R}^d$.  It is locally characterized by a coordinate chart, which provides a localized representation of $\calM$ as a subset of $\mathbb{R}^d$. For any specific point $\bp$ on the manifold $\calM$, a pertinent vector space, the tangent space $T_{\bp}\calM$, can be defined. This space is pivotal in elucidating the manifold's geometric characteristics. 
Moreover, a pseudo-Riemannian manifold is identified by an additional component, a pseudo-Riemannian metric $\bg$. For each point $\bp$ on $\calM$ and for any pair of vectors $\br, \bs$ in the tangent space $T_{\bp} \calM$ at $\bp$, an inner product $\langle{\br}, {\bs}\rangle_{\bg(\bp)}$ can be defined. 

A metric tensor (or simply metric) is an additional structure on a manifold $\calM$ that allows defining distances and angles, just as the inner product on a Euclidean space allows defining distances and angles there. 
The specific form of the metric depends on the geometry of the manifold. In Euclidean space, the metric is the identity matrix. In more complex spaces, such as curved manifolds, the metric can take on more intricate forms to account for the curvature of the manifold.

In the context of local chart coordinates, given $\bp= [p_1, \ldots, p_d] \T \in \calM, \br= [r_1, \ldots, r_d] \T \in T_{\bp}\calM$, and $\bs=[s_1, \ldots, s_d] \T \in T_{\bp}\calM$, the metric $\bg(\bp)$ is depicted as a real, symmetric, and non-degenerate (no zero eigenvalues) matrix. 
More precisely, a metric at a point $\bp$ of $\calM$ is a bilinear form defined on the tangent space $T_{\bp}\calM$ at $\bp$:
\begin{align}
\bg(\bp): T_{\bp} \mathcal{\calM} \times T_{\bp} \mathcal{\calM} \rightarrow \mathbb{R} .
\end{align}
The inner product with the metric can subsequently be computed as:
\begin{align}
\langle \br, \bs \rangle_{\bg(\bp)}:= \br\T \bg(\bp) \bs = \bs\T \bg(\bp) \br.
\end{align}

This assignment is expected to exhibit smooth variation with respect to the base point $\bp$ within $\calM$. The presence of this metric sets it apart from Euclidean space, which employs the Euclidean inner product as its metric.

\subsection{Neural Differential Diffusion}
\subsubsection{Neural Ordinary Differential Equations}
Ordinary differential equations (ODEs) are typically used to describe the dynamics of a system. The paper \cite{chen2018ode} introduces trainable neural ODEs, which parameterize the continuous dynamics of hidden units. The latent state of the ODE network, denoted as $\mby(t)$, is modeled as: 
\begin{align}
\ddfrac{\mby(t)}{t}=\calF^{\mathrm{ODE}}_{\theta}(\mby(t)), \label{eq:ode_f}
\end{align}
where $\calF^{\mathrm{ODE}}_{\theta}$ is a trainable network with weights $\theta$. 

Neural ODEs provide a flexible and adaptive framework for learning representations of data with the potential to capture high-order dependencies and intricate transformations. Recent research \cite{yan2019tisode,kang2021sodef} has also demonstrated that neural ODEs are inherently more resilient to input perturbations than regular neural networks. Moreover, neural ODEs \cite{chamberlain2021grand,chamberlain2021blend,zhao2023adversarial,she2023robustmat,wang2023robustloc} have been proposed and applied to GNNs to model the graph's diffusion process. In addition, the stability of the heat semigroup and the heat kernel under perturbations of the Laplace operator (i.e., local perturbation of the manifold) is studied in \cite{song2022robustness}.

\subsubsection{Neural Beltrami Diffusion}
Beltrami flow is a partial differential equation (PDE) widely used in hydrodynamics and geometric physics \cite{beltrami1,beltrami2,beltrami3,beltrami4}. 
The Beltrami diffusion on graphs is defined as \cite{song2022robustness}:
\begin{align}
\frac{ \mathrm{d} \bZ(\mu, t)}{\mathrm{d} t} = \frac{1}{2} \frac{1}{\norm{\nabla \bZ}} \mathrm{div} \parens*{ \frac{\nabla \bZ}{\norm{\nabla \bZ}}} (\mu, t), \label{eq.beltrami_diffusion}
\end{align}
where $\mathrm{div}$ denotes the divergence, $\nabla$ denotes the gradient operator, $\norm{}$ is a norm operator, $\bZ(\mu, t)$ denotes the features of the node with index $\mu$. 
By integrating the Beltrami flow into the graph neural network, the neural Beltrami diffusion can be constructed \cite{chamberlain2021blend}. 
In the neural Beltrami flow, the positional coordinates contribute to defining new graph topology with rewiring, which ensures more effective message passing. As the other advantage of neural diffusions with Beltrami flow, the robustness of feature representation for vertices is improved using both vertex features and positional features \cite{song2022robustness,chamberlain2021blend}. 



\subsection{Problem Formulation}
Given a query sensor frame, represented as a pair $(\bI, \bP)$, comprising an image frame $\bI$ and a point cloud frame $\bP$, our objective is to design a PR model, denoted as $f$, that effectively processes this multi-modal data, extracts features, and embeds the pair $(\bI,\bP)$ into a scene descriptor $\boldf$ of a significantly reduced dimension. The pipeline can be denoted as 
$f: (\bI,\bP) \mapsto \boldf$.
Subsequent to this embedding, a similarity assessment is performed between the query scene descriptor $\boldf$ and the set of all database scene descriptors. The query scene frame is deemed accurately localized if at least one of the top retrieved database scene frames lies within a pre-specified proximity to the ground truth position of the query. The processing pipeline is depicted in \cref{fig:pipeline}.


\section{PRFusion}\label{sec:prfuion}
\begin{figure*}[!htb]
\begin{center}
\includegraphics[width=1.\textwidth]{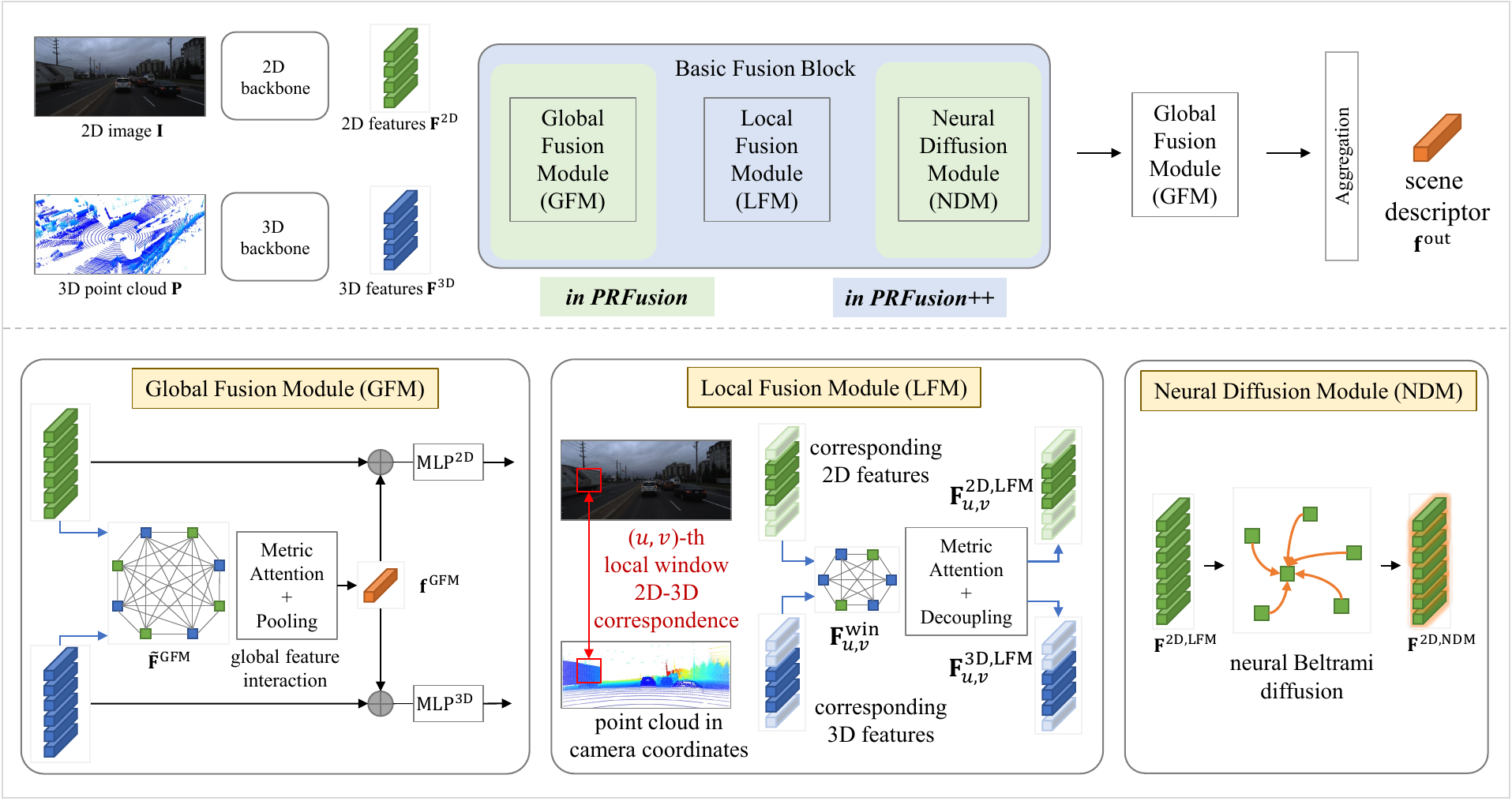}
\end{center}
\caption{The overall architecture of our proposed PRFusion and PRFusion++. The multi-modal fusion is conducted in both the GFM and the LFM. The image features are additionally passed through the NDM to enhance the feature robustness.}
\label{fig:architecture}
\end{figure*}

In this section, we provide details of the model PRFusion, which does not need camera-LiDAR extrinsic information. 
As depicted in \cref{fig:architecture}, given a sensor frame $(\bI, \bP)$, we initially employ two distinct preliminary backbone networks to separately extract 2D and 3D feature maps, denoted as $\bF^{\mathrm{2D}}\in\mathbb{R}^{hw\times c}$ and $\bF^{\mathrm{3D}}\in\mathbb{R}^{n\times c}$, respectively. Here, $h$ and $w$ represent the height and width of the 2D feature map, $n$ is the number of 3D features, and $c$ is the dimension of the feature\footnote{For illustration simplicity, we assume 2D and 3D features have the same dimension $c$.}. These different modal features are subsequently fed into the ensuing fusion blocks to facilitate feature interaction.

\subsection{Global Fusion Module (GFM)}\label{sec:global_fusion_module}
\subsubsection{Fusion Feature Initialization}
For multi-modal interaction, a pipeline that can effectively exploit information from distinct modalities is pivotal to the construction of the final scene descriptor $\boldf$. Nonetheless, previous multi-modal PR works \cite{komorowski2021minkloc++,lai2022adafusion} only consider fusing multi-modal features at the feature descriptor level, with interaction being applied only after the pooling operation. Such a late-stage processing strategy inherently overlooks a substantial amount of detailed information, making the multi-modal interaction less effective. To address this limitation, we introduce the GFM to achieve comprehensive 2D and 3D scene understanding.

To facilitate global multi-modal learning and generate a summarized vector to guide subsequent features, we construct a global multi-modal complete graph that comprises both 2D and 3D features. 
Given the significantly larger number of features, directly incorporating the entire features would impose a computational burden. Moreover, each feature, post backbone feature extraction, has already captured essential contextual information, thereby adequately representing the whole scene. 
Consequently, we opt for uniformly downsampled 2D and 3D features ${\bF^{\mathrm{2D}}}^{\prime} \in\mathbb{R}^{n^{\mathrm{2D}}\times c}$, ${\bF^{\mathrm{3D}}}^{\prime} \in\mathbb{R}^{n^{\mathrm{3D}}\times c}$ with  highly summarized pooling features to expedite multi-modal  interaction. Specifically, the initial graph features $\tilde{\bF}^{\mathrm{GFM}}  \in \mathbb{R}^{ \Tilde{n} \times c}$ with $\Tilde{n} = n^{\mathrm{2D}} + n^{\mathrm{3D}} + 2$ can be obtained as:
\begin{align}
\tilde{\bF}^{\mathrm{GFM}} = \set*{ {\bF^{\mathrm{2D}}}^{\prime}, {\bF^{\mathrm{3D}}}^{\prime}, \mathrm{Pooling}\parens*{\mathbf{F}^{\mathrm{2D}}},  \mathrm{Pooling}\parens*{\mathbf{F}^{\mathrm{3D}}} } , 
\end{align}
where $\mathrm{Pooling}\parens*{}$ is the global average pooling operation.

\subsubsection{Manifold Metric Attention}

The attention mechanism has demonstrated its strong capability in various fields \cite{vaswani2017transformer,dosovitskiy2020vit,lai2023spherical}. In multi-modal learning, a significant challenge arises from the diverse nature of feature sources and their embedding within different feature spaces. A static fusion space may be insufficient to accommodate the increased diversity of these features. To achieve more effective feature interaction, we propose to endow the multi-modal fusion space with an adaptable similarity measurement.

Typically, one constructs the attention weight by computing the inner products $\langle \bQ,\bK \rangle$ between the query features $\bQ$ and the key features $\bK$. In this work, we introduce a generalized manifold metric attention. 
Given $N$ points on a $c$-dimensional manifold, denoted as $\bp^{\calM}=[\bp_i^{\calM}]_{i=1}^N$, we establish the real symmetric non-degenerate metric at point $\bp_i^{\calM}$ as $\bg_i = \bg(\bp_i^{\calM})\in \Real^{c\times c}$.

Each  $\bQ_i$ and $\bK_j$ is interpreted as tangent vectors over a flexible manifold with a learnable local metric $ \bg_i $. This representation extends beyond vanilla attention, in which $\bg_i$ is a fixed identity matrix for each point $\bp_i^{\calM}$.

The multi-modal features, denoted as $\bp^{\calM}=\tilde{\bF}^{\mathrm{GFM}}$, are perceived as distinct base points within a manifold. These features construct their unique metric through a learnable neural ODE mapping defined as:
\begin{align} \label{eq:gfm_ode}
\frac{\mathrm{d} \bg_i(t)}{\mathrm{d} t} = \sigma( \bg_i(t) \bW_{\bg}),
\end{align}
where $\bg_i(0) =  \tilde{\bF}^{\mathrm{GFM}}_i$, $\sigma \parens{}$ is an non-linear activation function, and $\bW_{\bg}\in \Real^{c\times c}$ is a learnable matrix. 
By solving \cref{eq:gfm_ode}, we obtain the diffused manifold metric at each base point as $\bG_i$.
Neural ODEs provide a flexible and adaptive framework for learning representations and are able to capture high-order dependencies, which supports constructing a more adaptive manifold metric and contributes to more fine-grained feature interaction of different modalities.

Each of the weighted global fusion features $\bF_i^{\mathrm{GFM}} \in \Real^{c}$ is then obtained as follows:
\begin{align}
\bF_i^{\mathrm{GFM}}  &=  \sum_j a_{i,j} \tilde{\bF}_j^{\mathrm{GFM}} \bW_{\bV}, \label{eq:F_gfm} \\
a_{i,j} &=  \mathrm{Softmax} \parens*{ \left\langle \tilde{\bF}_i^{\mathrm{GFM}} \bW_{\bQ} , \tilde{\bF}_j^{ \mathrm{GFM}}\bW_{\bK}\right\rangle_{\mathrm{diag} (\bG_i)  }}, \label{eq:aij}
\end{align}
where $\mathrm{Softmax}\parens*{}$ denotes the row-wise softmax operation, $\bW_{\bQ}, \bW_{\bK}, \bW_{\bV}$ are learnable matrices, and $\mathrm{diag}\parens*{}$ is the vector-to-matrix diagonal operation.

\subsubsection{Feature Updating}
The global representation of the scene can guide the 2D and 3D modality features towards better representation learning. 
To this end, we aggregate the whole global feature as a highly condensed feature vector $\mathbf{f}^{\mathrm{GFM}} \in \Real^{c}$ summarized by the GeM pooling \cite{radenovic2018gempooling}, $\calF_{p}\parens*{}$, with a learnable parameter $p$ defined as: 
\begin{align}\label{eq:gem} 
\mathbf{f}^{\mathrm{GFM}} = \calF_{p}\parens*{\bF^{\mathrm{GFM}}} \coloneqq \parens*{\frac{1}{ \Tilde{n} } \sum_{i=1}^{ \Tilde{n} } \parens*{\bF^{\mathrm{GFM}}_{i}}^{p}}^{\frac{1}{p}}, 
\end{align}
where $(\cdot)^p$ denotes raising each element to the power of $p$.
We employ this vector as an element-wise additive bias to compute the biased 2D and 3D feature maps, enabling the respective modality features to encode holistic scene representations. The biased feature maps are then passed through separate MLPs to obtain the respective updated 2D or 3D feature maps, denoted as: 

\begin{align} 
\bF^{\mathrm{2D,GFM}} &= \mathrm{MLP}^{\mathrm{2D}} \parens*{ \bF^{\mathrm{2D}} \oplus \mathbf{f}^{\mathrm{GFM}}} \in \bbR^{hw\times c}, \label{eq:mlp2d} \\ 
\bF^{\mathrm{3D,GFM}} &= \mathrm{MLP}^{\mathrm{3D}} \parens*{ \bF^{\mathrm{3D}} \oplus \mathbf{f}^{\mathrm{GFM}}} \in \bbR^{n\times c}, \label{eq:mlp3d}
\end{align}
where $\oplus$ denotes broadcast addition.

\subsection{Neural Diffusion Module (NDM)}
In challenging scenes where there are environmental perturbations, the sensors can be easily affected, leading to inferior feature extraction and poor PR performance. Image-based models are more susceptible to challenging environmental conditions than point cloud-based ones (see \cref{tab:boreas}). Therefore, to achieve more robust feature expression, we resort to leveraging neural diffusion layers, where the outputs $\bF^{\mathrm{2D,GFM}}$ from the GFM are additionally passed into the NDM for feature enhancement.

Specifically, we employ a neural diffusion mechanism rooted in the Beltrami flow. This mechanism is applied to each feature denoted as $\bX(t)$, with $\bX(0)=\bF^{\mathrm{2D,GFM}}$. The  neural diffusion module is characterized as:
\begin{align}
\frac{ \mathrm{d} \bX(t) }{\mathrm{d} t} &= \overline{A} \bX(t) \bW_{\bX} =(\bA(t)-\bE )    \bX(t) \bW_{\bX},  \label{eq:ndm} \\
\bA(t) &= \mathrm{Softmax} \parens*{ \mathrm{KNN} \parens*{\bY(t) \bY\T(t)  }},  \\
\bY(t) &= \bX(t) \bW_{\bY},
\end{align}
where $\bE$ is the identity matrix, $\mathrm{KNN}\parens*{}$ is the $K$-nearest neighbor algorithm, $\bW_{\bX}, \bW_{\bY}$ are learnable parameters. The neural Beltrami diffusion can be regarded as a type of graph neural diffusion, which is additionally equipped with graph topology rewiring \cite{chamberlain2021blend} achieved by positional embeddings. Graph rewiring enables the construction of flexible graph node connections and can thus contribute to better node feature learning.

By solving the above neural Beltrami diffusion equations, we obtain the 2D Beltrami features as $\bF^{\mathrm{2D,NDM}}$. Subsequently, the 2D output $\bF^{\mathrm{2D,NDM}}$ from the NDM and the 3D output $\bF^{\mathrm{3D,GFM}}$ from the last GFM are together fed into the final GFM to achieve cascaded feature updating.

\subsection{Output Scene Descriptor Generation}
In the PRFusion framework, as shown in \cref{fig:architecture}, we stack the above-described GFM for 2 layers with 1 intermediate NDM layer to achieve cascaded feature fusion. The final scene descriptor is obtained by concatenating the fusion, 2D, and 3D feature vectors obtained from the final GFM in \cref{eq:gem,eq:mlp2d,eq:mlp3d}:
\begin{align} \label{eq:output}
\mathbf{f}^{\mathrm{out}} = \mathbf{f}^{\mathrm{GFM}}  \concat \calF_{p^{\mathrm{2D}}}\parens*{\bF^{\mathrm{2D,GFM}}}  \concat  \calF_{p^{\mathrm{3D}}}\parens*{  \bF^{\mathrm{3D,GFM}} },
\end{align}
where $p^{\mathrm{2D}}$ and $p^{\mathrm{3D}}$ are learnable parameters for the GeM pooling $\calF_{p}\parens*{}$ as in \cref{eq:gem}.

\section{PRFusion++}\label{sec:prfusion++}
The PRFusion model, as described above, is designed for multi-modal PR in scenarios where the extrinsic parameters between the camera and LiDAR sensors are unknown. Without knowing the extrinsic calibration, one cannot establish the correspondences between image pixels and point clouds. However, in most applications, the sensor extrinsic parameters can be determined during the sensor setup stage. With this additional information, we can conduct multi-modal feature interaction at a more fine-grained level. Consequently, we propose an enhanced version of PRFusion, namely PRFusion++, which can leverage detailed 2D/3D information, as depicted in \cref{fig:architecture}. 

\subsection{Local Fusion Module (LFM)}\label{sec:local_fusion_module}
We denote the real-world positions of the 3D point cloud features $\bF^{\mathrm{3D,GFM}}$ (from GFM in \cref{eq:mlp3d}) as $[\bx, \by, \bz] \in \mathbb{R}^{n\times 3}$. They are the input $3$-dimensional point cloud coordinates. 
The projected points $\bI^{\mathrm{3D}} = [\bu^{\mathrm{3D}}, \bv^{\mathrm{3D}}] \in \mathbb{R}^{n\times 2}$ on the image feature plane, derived from 3D cloud points, are given by the camera-LiDAR projection geometry.

Similarly, we can denote the image feature plane positions of the output 2D pixels as $\bI^{\mathrm{2D}} = [\bu^{\mathrm{2D}}, \bv^{\mathrm{2D}}]  \in \mathbb{R}^{hw\times 2}$.

The point cloud projection geometry enables an association between 2D and 3D features on the image plane, as illustrated in \cref{fig:architecture}. This association facilitates the definition of a neighboring relation between 2D and 3D feature nodes. 
Hence, we introduce the LFM to construct local corresponding feature learning. Specifically, we divide the entire image feature frame into distinct non-overlapping windows \cite{dosovitskiy2020vit,liu2021swin,lai2023spherical}, each with a size of $\Delta h \times \Delta w$. Each $(u,v)$-th window graph can contain a varying number of 2D and 3D feature nodes:
\begin{align}\label{eq:local_window_partition}
\bF^{\mathrm{win}}_{u,v} = \
&\ml{\set*{\bF^{\mathrm{2D,GFM}}_j           \given   \floor*{\frac{\bI^{\mathrm{2D}}_j}{[\Delta h, \Delta w]}}             = [u,v]   }_{j\in [hw]}  \\
\bigcup \set*{\bF^{\mathrm{3D,GFM}}_{j^{\prime}} \given   \floor*{\frac{\bI^{\mathrm{3D}}_{j^{\prime}}}{[\Delta h, \Delta w]}} = [u,v]   }_{j^{\prime}\in [n]},
}
\end{align}
where $u \in \set*{1, \ldots, \floor*{\frac{h}{\Delta h}} }$ and $v \in \set*{1, \ldots, \floor*{\frac{w}{\Delta w}}}$. Within each window, we establish complete connections among all feature nodes, enabling each node to pass feature messages to all other nodes within the same window.

We then employ the manifold metric attention defined from \cref{eq:gfm_ode,eq:F_gfm,eq:aij} to perform local multi-modal feature updating for each window to obtain the updated features $\bF^{\mathrm{LFM}}_{u,v}$ based on $\bF^{\mathrm{win}}_{u,v}$. Note that the attention computation in all windows is proceeding in parallel with the assistance of the varied-length dot product package \cite{lai2023spherical}.
By decoupling the mixed feature nodes, $\bF^{\mathrm{2D,LFM}}_{u,v}$ and $\bF^{\mathrm{3D,LFM}}_{u,v}$, from each window $\bF^{\mathrm{LFM}}_{u,v}$, we obtain the updated 2D and 3D features respectively as:
\begin{align} \label{eq:2dlfm}
\bF^{\mathrm{2D,LFM}} &= \concat_{u,v} {\bF^{\mathrm{2D,LFM}}_{u,v}},  \\
\bF^{\mathrm{3D,LFM}} &= \concat_{u,v} {\bF^{\mathrm{3D,LFM}}_{u,v}}, \label{eq:3dlfm}
\end{align}
where the updated 2D features are fed into the NDM for robustness enhancement as shown in \cref{fig:architecture}.

\subsection{Output Scene Descriptor Generation}
Different from PRFusion, besides 2 GFMs and 1 NDM, PRFusion++ is additionally equipped with 1 LFM as shown in \cref{fig:architecture}. The final output scene descriptor is obtained using the outputs from the final GFM as in \cref{eq:output}.

\section{Loss Function}
Both of our proposed models, PRFusion and PRFusion++, adopt the triplet loss \cite{hoffer2015tripletloss}:
\begin{align}
\ell = \mathrm{max}\parens*{ \
\norm{\mathbf{f}^{\mathrm{a}} - \mathbf{f}^{\mathrm{p}} }_{2} - \
\norm{\mathbf{f}^{\mathrm{a}} - \mathbf{f}^{\mathrm{n}} }_{2} + m, 0 },
\end{align}
where $\mathbf{f}^{\mathrm{a}}, \mathbf{f}^{\mathrm{p}}, \mathbf{f}^{\mathrm{n}}$ are descriptors of an anchor, a positive sample and a negative sample, and $m$ is the margin hyperparameter.

\section{Experiments}\label{sec:experiments}
In this section, we compare our proposed models with other baselines in different datasets. We also conduct necessary ablation studies to demonstrate the effectiveness of our proposed modules.

\subsection{Implementation Details and Datasets}

\subsubsection{Implementation Details} 
Our image and point cloud backbones are constructed based on MinkLoc++ \cite{komorowski2021minkloc++}. We set the maximum batch size as $160$. We train our network for a total of $120$ epochs. The Adam optimizer \cite{kingma2014adam} with a maximum learning rate $1e-3$ and weight decay $1e-4$ is used to train the network. We use SpTr \cite{lai2023spherical} for varied-length multi-modal attention computing. The package torchdiffeq\cite{chen2018ode} is used for neural ODE solving. The input image is resized such that the short side is $240$. The voxel quantization method is applied to input point clouds, where the quantization size is set as $0.1$ for the Oxford dataset and $1$ for the Boreas and KITTI datasets. Necessary data augmentation techniques are applied during training. The experiments are conducted on a Tesla A100.

\subsubsection{Evaluation Metrics}
We follow previous works to use the same evaluation protocol for PR, including Recall@1 (R@1), Average Recall@$N$ (AR@$N$), and Average Recall@1\% (AR@1\%). Unless otherwise noted, we set the positive retrieval threshold as $25$ m, i.e., if a retrieval is within $25$ m from the query ground truth position, this retrieval is treated as a positive (successful) retrieval.

\subsubsection{Datasets}
The Oxford RobotCar dataset \cite{oxford} is a large-scale autonomous driving dataset, including a wide range of driving conditions. We use the processed point clouds provided by PointNetVLAD\cite{uy2018pointnetvlad} which is the standard benchmark data for point cloud PR. We test on both the standard benchmark split (denoted as Oxford-PNVLAD) and the split used by \cite{oertel2020cues,lai2022adafusion} (denoted as Oxford-Cues). Since the processed point clouds break the camera-LiDAR extrinsic, we can only test PRFusion, which does not require the camera-LiDAR alignment.

\begin{figure*}[!htb]
\begin{center}
\includegraphics[width=0.9\textwidth]{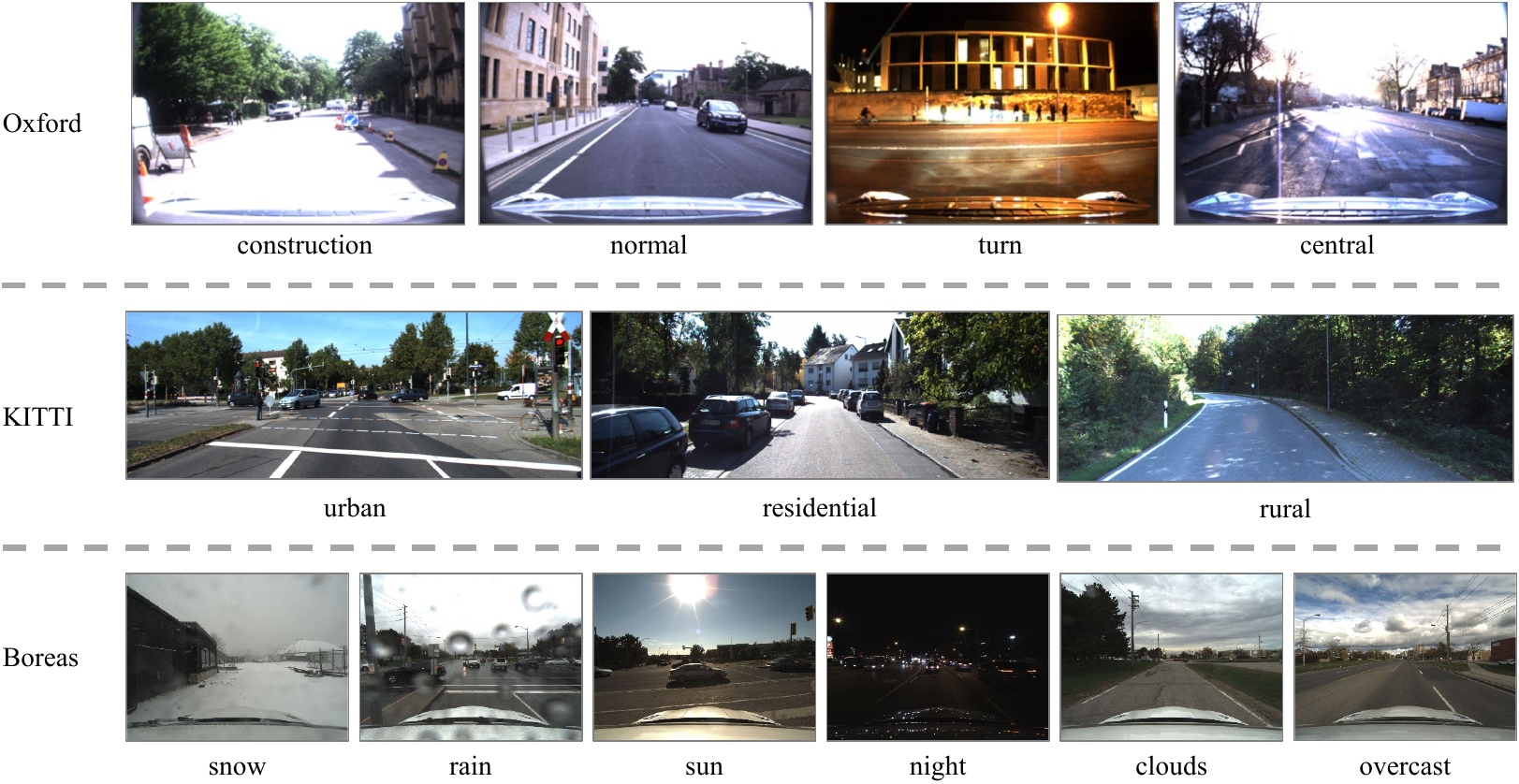}
\end{center}
\caption{Examples from the Oxford, KITTI, and Boreas datasets.}
\label{fig:dataset_viz}
\end{figure*}

We also test on the KITTI dataset\cite{geiger2013kitti}, which is widely used in computer vision and autonomous driving research as it offers a diverse range of real-world autonomous vehicle data in different environments. It has provided camera-LiDAR extrinsic calibrations.

Both of the above two datasets are collected in normal conditions, which may not be sufficient to demonstrate the model's robustness. The Boreas dataset \cite{burnett2022boreas} is a more challenging dataset collected in various extreme conditions including snow, rain, and night. It also provides calibrated images and point clouds with synchronized extrinsic parameters. 

Visual examples from the above datasets are provided in \cref{fig:dataset_viz}.

\subsection{Main Results}
We begin by comparing PRFusion with other baselines on the Oxford dataset. As shown in \cref{tab:oxford-pnvlad,tab:oxford-cues}, PRFusion achieves state-of-the-art performance in all metrics. This demonstrates the effectiveness of PRFusion. Furthermore, multi-modal PR approaches that take point clouds and images as inputs tend to be more accurate than those using only one modality. This verifies that images captured by cameras can provide additional useful information for the PR task.

\begin{table}[!hbt]
\centering
\caption{PR results on the Oxford-PNVLAD dataset. All models do not use the re-ranking technique. "-" denotes that the result is not provided by the corresponding paper. "*" denotes that the model is trained with extra datasets. The best and the second best performances are marked with \first{bold} and \second{underline}, respectively. }\label{tab:oxford-pnvlad}
\begin{tabular}{c  l  |  c   c    } 
\toprule
& \multirow{2}{*}{Model}  &  \multicolumn{2}{c}{Oxford-PNVLAD}\\
\cmidrule(lr){3-4} & & AR@1\%  &  AR@1   \\ 
\midrule
\multirow{8}{*}{3D} & PointNetVLAD \cite{uy2018pointnetvlad}   & 80.3   & 63.3    \\
& PCAN \cite{zhang2019pcan}                & 83.8   & 70.7    \\
& LPD-Net \cite{2019lpdnet}                    & 94.9   & 86.4    \\
& EPC-Net \cite{2022epcnet}                    & 94.7   & 86.2    \\
& SOE-Net \cite{2021soenet}                    & 96.4   & 89.3    \\
& MinkLoc3D \cite{komorowski2021minkloc3d} & 97.9   & 93.8    \\
& MinkLoc3Dv2 \cite{minkloc3dv2}           & 98.9   & 96.3    \\
& PTC-Net \cite{chen2023ptcnet}            & 98.8   & 96.4    \\
\midrule
\multirow{6}{*}{\makecell[c]{3D\\+\\RGB}} & CORAL \cite{pan2021coral}                & 96.1  &  -  \\
& PIC-Net \cite{lu2020picnet}              & 98.2  &  -  \\
& MinkLoc++ \cite{komorowski2021minkloc++} & 99.1  & 96.7 \\
& UMF\cite{garcia2024umf}   &  \second{99.2}  &  97.2 \\
& *UMF\cite{garcia2024umf} w/ extra train data    &  99.1  &  \second{97.9} \\
& PRFusion (ours) & \first{99.6}   &  \first{98.2}   \\
\bottomrule
\end{tabular}
\end{table}

\begin{table}[!hbt]
\centering
\caption{PR results on the Oxford-Cues dataset. All models do not use the re-ranking technique or additional training datasets. "-" denotes that the result is not provided by the corresponding paper. The best and the second best performances are marked with \first{bold} and \second{underline}, respectively.}
\label{tab:oxford-cues}
\begin{tabular}{c  l   |  c   c   } 
\toprule
 & \multirow{2}{*}{Model}   &  \multicolumn{2}{c}{Oxford-Cues}  \\
\cmidrule(lr){3-4} &  & AR@1\%  &  AR@1   \\ 
\midrule
\multirow{3}{*}{\makecell[c]{3D\\+\\RGB}}& Cues-Net \cite{oertel2020cues}           & -     & 98.00 \\
& AdaFusion \cite{lai2022adafusion}        & \second{99.21}  &  \second{98.18}  \\
& PRFusion (ours) & \first{99.94}  &  \first{99.08}  \\
\bottomrule
\end{tabular}
\end{table}

\begin{figure}[!htb]
\centering
\includegraphics[width=0.8\columnwidth]{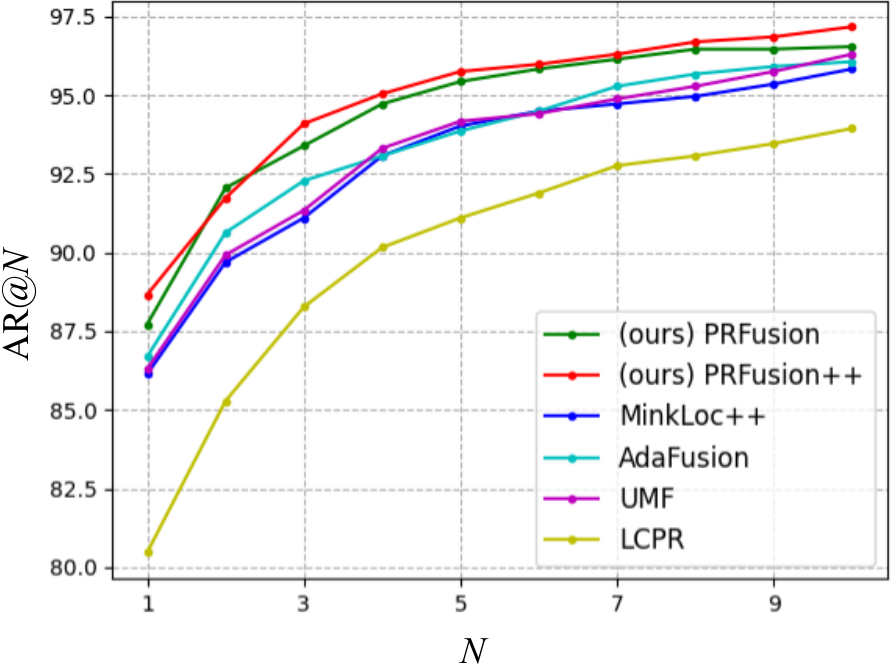}
\caption{Average Recall@$N$ curve on the KITTI dataset.}
\label{fig:arn_kitti}
\end{figure}

We next conduct experiments on the KITTI dataset that encompasses various typical driving scenarios. As illustrated in \cref{tab:kitti}, our proposed models consistently outperform other baselines across different evaluation metrics. The corresponding Average Recall@$N$ curve, depicted in \cref{fig:arn_kitti}, demonstrates the superiority of our models. 

\begin{figure}[!htb]
\centering
\includegraphics[width=0.8\columnwidth]{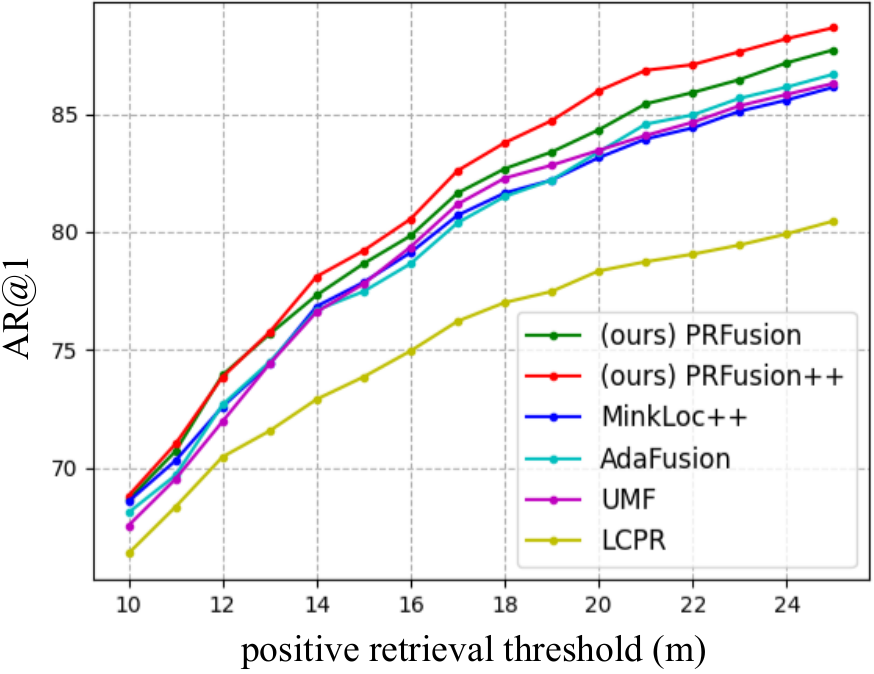}
\caption{AR@1 under different positive retrieval thresholds on the KITTI dataset.}
\label{fig:thrshold_kitti}
\end{figure}

We also visualize AR@1 under different positive retrieval thresholds (from $10$ m to $25$ m) as in \cref{fig:thrshold_kitti}. In challenging small thresholds ($<16$ m), comparable performance is observed with other baselines. However, under larger thresholds ($>17$ m), we achieve significantly better AR@1.

\begin{table}[!htb]
\centering
\caption{PR results on the KITTI dataset. The best and the second best performances are marked with \first{bold} and \second{underline}, respectively.}
\label{tab:kitti}
\begin{tabular}{c l  |   c  c  c   c} 
\toprule
& Model      &   AR@1  &  AR@2  &  AR@5  &  AR@10  \\
\midrule
\multirow{6}{*}{\makecell[c]{3D\\+\\RGB}} &  MinkLoc++\cite{komorowski2021minkloc++}  &   86.1   &  89.7  &   94.0   &  95.8  \\
&  AdaFusion\cite{lai2022adafusion}         &   86.7   &  90.6  &   93.9   &  96.1  \\
&  UMF\cite{garcia2024umf}                  &   86.3   &  89.9  &   94.2   &  96.3   \\
&  LCPR\cite{zhou2023lcpr}                  &   80.5   &  85.3  &   91.1   &  93.9   \\
&  PRFusion (ours)                          &   \second{87.7}   &  \first{92.1}  &   \second{95.4}   &  \second{96.5}   \\
&  PRFusion++ (ours)                        &   \first{88.7}   &  \second{91.7}  &   \first{95.8}   &  \first{97.2}   \\
\bottomrule
\end{tabular}
\end{table}

Next, we evaluate our models on the challenging Boreas dataset, which includes camera-LiDAR extrinsic calibrations and demonstrates the strong capacity of our advanced model, PRFusion++. As illustrated in \cref{tab:boreas}, PRFusion++ achieves the highest score in all metrics. 
The superiority of leveraging multi-modal data is also evident in \cref{tab:boreas}, as all multi-modal models outperform the single-modal ones. We also note that the point cloud models generally outperform the image models, particularly in the night scenario with extremely poor lighting conditions, where all image models exhibit significant performance drops while the point cloud models remain almost unchanged.

\begin{table*}[!htb]\footnotesize
\centering
\caption{PR results on the Boreas dataset. All models do not use the re-ranking technique or additional training datasets. The best and the second best performances are marked with \first{bold} and \second{underline}, respectively.}
\label{tab:boreas}
\begin{tabular}{ c   l   |   c   c   c   c   c   c    c   c } 
\toprule
&\multirow{2}{*}{Model}  &    Snow  & Rain  & Sun  & Night  & Clouds  &  Overcast & \multirow{2}{*}{AR@1 \%}  & \multirow{2}{*}{AR@1}  \\
\cmidrule(lr){3-8} & & R@1  &  R@1   &  R@1  &  R@1   &  R@1  &  R@1  &   \\ 
\midrule
\multirow{3}{*}{RGB}&GeM \cite{radenovic2018gempooling}     &  62.9  &   69.1  &   67.1  &  18.6  &   69.5  &  68.5   &   76.2   &  59.3 \\
& ConvAP \cite{ali2022convap}                               &  75.4  &   74.6  &   72.7  &  24.8  &   75.0  &  75.1   &   79.4   &  66.3 \\
& MixVPR \cite{ali2023mixvpr}                               &  81.3  &   80.3  &   78.3  &  30.2  &   80.0  &  81.0   &   85.5   &  71.9 \\
\midrule
\multirow{3}{*}{3D} & MinkLoc3D \cite{komorowski2021minkloc3d}   &  77.0  &   89.5  &   90.0  &  89.3  &   88.6  &  89.6  &   96.6   &  87.3  \\
& MinkLoc3Dv2 \cite{minkloc3dv2}                                 &  84.1  &   92.6  &   92.6  &  92.3  &   93.1  &  92.9  &   98.2   &   91.3 \\
& PTC-Net \cite{chen2023ptcnet}                                  &  82.2  &   91.4  &   92.2  &  92.0  &   92.8  &  93.1  &   97.4   &   90.6 \\
\midrule
\multirow{5}{*}{3D + RGB}& MinkLoc++ \cite{komorowski2021minkloc++} &  87.1   &  93.3  &  94.6  &  90.9  &  94.5  &  94.5  &  98.9  &  92.5 \\
& AdaFusion \cite{lai2022adafusion}                                 &  88.1   &  93.5  &  94.8  &  91.5  &  94.8  &  94.7  &  99.0  &  92.9 \\
& UMF \cite{garcia2024umf}                                          &  87.5   &  94.2  &  94.0  &  91.4  &  93.8  &  93.4  &  99.0  &  92.4 \\
& LCPR\cite{zhou2023lcpr}  & 70.3  &  90.2  &  88.9  &  85.1  & 89.6  & 89.7  &  95.3   &  85.6 \\
& PRFusion (ours)   &  \second{92.0}    &  \second{96.4}    &  \second{95.0}    &   \second{92.5}    &  \second{95.9}   &  \second{95.8}    &   \second{99.2}   &   \second{94.6}  \\
& PRFusion++ (ours) &  \first{94.2}    &  \first{96.7}    &  \first{96.1}    &   \first{95.9}    &  \first{96.1}   &  \first{96.3}    &   \first{99.6}   &   \first{95.9}  \\
\bottomrule
\end{tabular}
\end{table*}

\subsection{Design Analysis}\label{sec:design_analysis}
To verify the effectiveness of the proposed modules, we conduct design analysis for PRFusion++ in the challenging Boreas dataset.

\subsubsection{Module Ablation Study}
We first ablate the proposed modules from PRFusion and PRFusion++ as shown in \cref{tab:ablation}, where they all contribute to better performance. The three proposed modules (GFM, NDM, and LFM) together yield +3.4 gains for AR@1, underscoring their efficacy in enhancing overall outcomes. Among them, the GFM and LFM contribute more than the NDM.


\begin{table}[!htb]
\centering
\caption{Main ablation study on the proposed modules.}
\label{tab:ablation}
\begin{tabular}{ l  |   c  c} 
\toprule
Method      &  AR@1  &  $\Delta$\\
\midrule
PRFusion      &  \textbf{94.6}  &  - \\
PRFusion w/o  global fusion module (GFM)     &  93.2  &  -1.4 \\
PRFusion w/o  neural diffusion module (NDM)  &  93.8  &  -0.8 \\
w/o GFM/NDM  &  92.5  &  -2.1 \\
\midrule
PRFusion++       &  \textbf{95.9}  &  - \\
PRFusion++ w/o global fusion module (GFM)            &  94.7  & -1.2 \\
PRFusion++ w/o neural diffusion module (NDM)         &  95.3  & -0.6 \\
PRFusion++ w/o local fusion module (LFM)             &  94.6  & -1.3 \\
w/o GFM/NDM/LFM  & 92.5 & -3.4\\
\bottomrule
\end{tabular}
\end{table}

\subsubsection{Manifold Metric Attention}

In \cref{tab:metric_attention}, we present a comparative analysis among MLP, the conventional vanilla attention mechanism, and our proposed metric attention approach.
Our metric attention scheme yields a noteworthy improvement of +2.2 in the AR@1 metric. Compared to vanilla attention, which contributes +1.5, our metric attention demonstrates a +0.7 better performance. This indicates its efficacy in facilitating more robust multi-modal interactions through the acquired adaptive manifold metric.
Furthermore, we conduct ablation experiments on the metric attention module by selectively removing the neural ODE and the non-linear activation function. These ablations result in a noticeable performance degradation, underscoring the critical role played by the amalgamation of higher-order ODE modeling and non-linear activation functions in the construction of our flexible metric framework.

\begin{table}[!htb]
\centering
\caption{Comparison of different types of vanilla attention and our proposed metric attention.}
\label{tab:metric_attention}
\begin{tabular}{ l  |   c  c} 
\toprule
Method    &  AR@1  & $\Delta$\\
\midrule
w/o attention (MLP)                      &  93.7    & -   \\
vanilla attention                        &  95.2    & +1.5   \\
metric attention w/o ODE                 &  95.5    & +1.8   \\
metric attention w/o activation          &  95.6    & +1.9   \\
metric attention                         &  \first{95.9}  & \first{+2.2}   \\
\bottomrule
\end{tabular}
\end{table}

\subsubsection{Sampling Points in the GFM}

Within the GFM, we adopt a strategy of leveraging down-sampled 2D/3D features for the synthesis of multi-modal features. As in \cref{tab:sample}, with a modest number of sampled features, we are able to effectively encapsulate global-level representations. This empirical observation serves as compelling evidence that our design is highly effective and well-suited for its intended purpose.

\begin{table}[!htb]
\centering
\caption{Comparison on the number of sampled points in the GFM.}
\label{tab:sample}
\begin{tabular}{ l  |   c} 
\toprule
\#Points              &  AR@1 \\
\midrule
8                     &  95.2 \\
16                    &  \first{95.9} \\
32                    &  95.7 \\
64                    &  95.8 \\
\bottomrule
\end{tabular}
\end{table}

\subsubsection{Fusion Window Size in the LFM}
In the LFM, we perform 2D/3D feature interaction within a local window. Our experimentation, as outlined in \cref{tab:window}, involves the exploration of varying window sizes. Intriguingly, the results clearly demonstrate that the smallest window size, 1$\times$1, emerges as the optimal selection. This is attributed to its capacity to facilitate fine-grained local feature interactions.
Conversely, with the increasing window size, the performance decreases steadily. This indicates that trivially conducting full-scale multi-modal feature interaction is not a suitable solution for feature updating.

\begin{table}[!htb]
\centering
\caption{Comparison on different window size $\Delta h \times \Delta w$  in the LFM.}
\label{tab:window}
\begin{tabular}{ l  |    c} 
\toprule
$\Delta h \times \Delta w$             &  AR@1 \\
\midrule
1$\times$1                      &  \first{95.9} \\
2$\times$2                      &  95.2 \\
4$\times$4                      &  94.3 \\
8$\times$8                      &  92.1 \\
\bottomrule
\end{tabular}
\end{table}

\subsubsection{Neighborhood Scale in the NDM}
The neural Beltrami diffusion in the NDM is based on adaptive neighborhood construction, and we test the performance with different neighbors in the NDM. As illustrated in \cref{tab:neighbor}, a medium scale with 25 neighbors is an optimal choice for Beltrami feature diffusion. Further bringing more neighbors can not contribute to better performance.

\begin{table}[!htb]
\centering
\caption{Comparison on the number of nearest neighbors in the NDM.}
\label{tab:neighbor}
\begin{tabular}{ l |    c} 
\toprule
\#Neighbors              &  AR@1 \\
\midrule
9                      &  95.4 \\
16                     &  95.7 \\
25                     &  \first{95.9} \\
36                     &  95.6 \\
\bottomrule
\end{tabular}
\end{table}

\subsubsection{Robustness Against Image Perturbations}\label{subsubsec:robustness_gaussian}
\begin{figure}[!htb]
\centering
\includegraphics[width=0.41\textwidth]{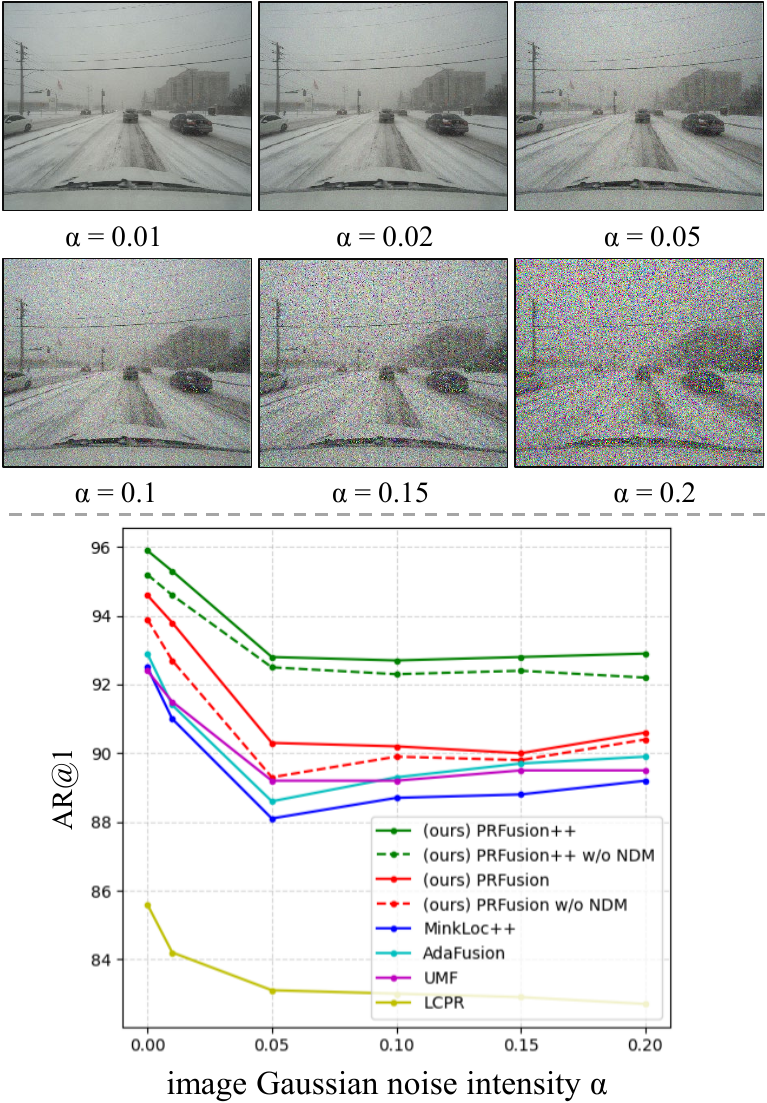}
\caption{Above: visualization of images under additive Gaussian noise with different noise intensities $\alpha$. Below: performance plot (AR@1) under additive image Gaussian noise with different noise intensities $\alpha$.}
\label{fig:draw_gaussian}
\end{figure}

\begin{figure}[!htb]
\centering
\includegraphics[width=0.47\textwidth]{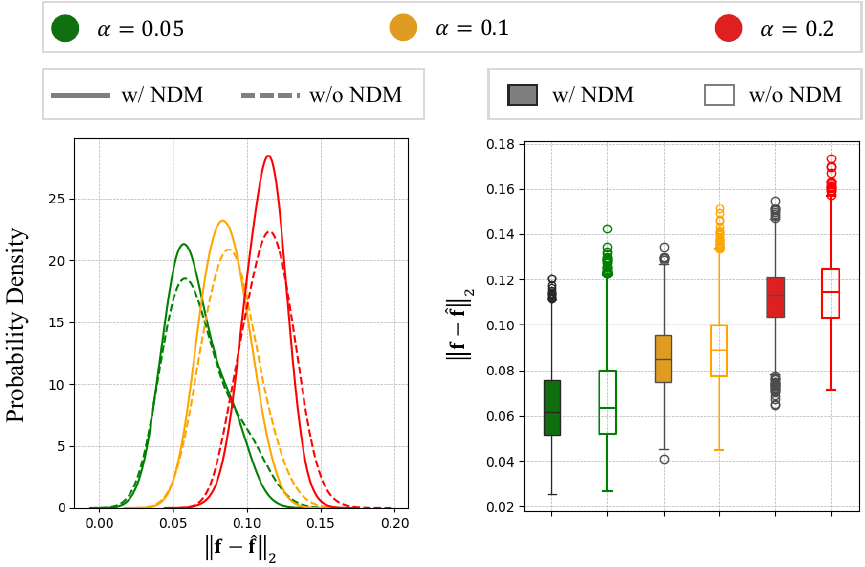}
\caption{Left: kernel density estimate plots. Right: box plots of $\norm{\mathbf{f}-\hat{\mathbf{f}}}_2$ under additive Gaussian noise with different noise intensities $\alpha$.}
\label{fig:draw_kde_boxplot}
\end{figure}

To verify the model robustness in more challenging scenarios, we test to add additional Gaussian noise onto the original image $\bI$. The perturbed image is built as $\hat{\bI} = \bI + \alpha \delta$, where $\delta \sim \calN(0,1)$ and $\alpha > 0$ controls the intensities, and the descriptor of the perturbed scene can be denoted as $\hat{\mathbf{f}}$. The visualization of perturbed images with different intensities is shown in \cref{fig:draw_gaussian}. As plotted in \cref{fig:draw_gaussian}, our proposed models PRFusion and PRFusion++ both show significantly better robustness against image perturbations, which strongly underscores the efficacy of our design. We also test the robustness by excluding the NDM, which leads to inferior performance in challenging scenarios. Moreover, it can be noticed from \cref{fig:draw_gaussian} that the LFM in PRFusion++ can boost model robustness for a large margin compared with PRFusion.
In addition, we visualize the kernel density estimate plots and box plots as in \cref{fig:draw_kde_boxplot}. With the integration of the NDM, our model can lead to smaller $\norm{\mathbf{f}-\hat{\mathbf{f}}}_2$ compared with that without the NDM. The robustness of the NDM can thus be demonstrated.


\subsubsection{Robustness Against Extrinsic Calibration Errors}\label{subsubsec:robustness_extrinsic}
\begin{figure}[!htb]
\centering
\includegraphics[width=0.8\columnwidth]{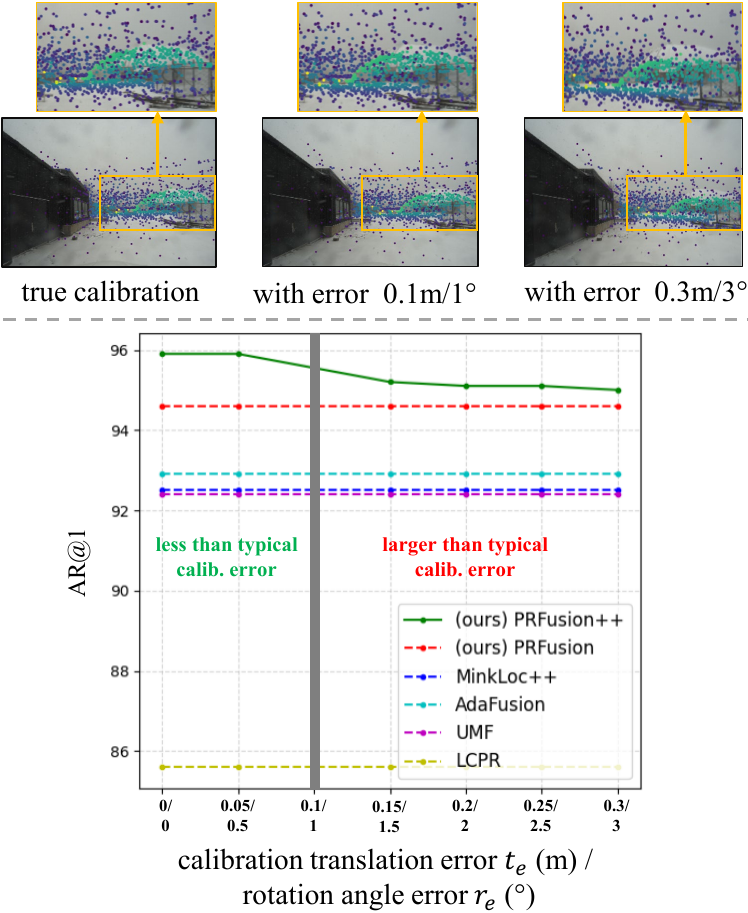}
\caption{Above: visualization of images and point clouds under camera-LiDAR extrinsic parameter calibration errors. Below: performance plot (AR@1) under camera-LiDAR extrinsic parameter calibration errors. Note that different from PRFusion++, PRFusion and other baselines are not affected by extrinsic parameter calibration errors and their performance plots are flat.}
\label{fig:draw_extrinsics}
\end{figure}

We evaluate the robustness against camera-LiDAR extrinsic calibration parameter errors. We denote the true calibration translation vector and calibration rotation matrix as $\bt$ and $\bR$ respectively. The measured calibration translation vector and rotation matrix can be denoted as $\hat{\bt} = \bt + \bt_{e}$ and $\hat{\bR}=\bR_{e}\bR$ with rotation matrix and translation vector errors $\bR_{e}$ and $\bt_{e}$. The calibration translation error $t_e$ and rotation angle error $r_e $  are given as:
\begin{align*}
t_e &= \norm{\bt_e}_2,\\
r_e &= \abs{ \mathrm{arccos}\parens*{\frac{\mathrm{trace} (\bR_e) -1 }{2}}} .
\end{align*}
The visualization of different extrinsic calibration errors is shown in \cref{fig:draw_extrinsics}. The performance comparison under different error thresholds is shown in \cref{fig:draw_extrinsics}. 
The results demonstrate that reasonably good extrinsic calibration is essential for the PRFusion++ model as it utilizes the camera-LiDAR extrinsic calibration information. In practice, the extrinsic calibration error is typically less than 0.1 m and 1$^{\circ}$\cite{zhang2004extrinsic,zhou2018automatic,schneider2017regnet,iyer2018calibnet,lv2021lccnet}. From \cref{fig:draw_extrinsics}, we see that under such errors, PRFusion++ outperforms other baselines, demonstrating its effectiveness.

\begin{table}[!htb]
\centering
\caption{Runtime speed and GPU memory usage on a Tesla A100.}
\label{tab:speed}
\begin{tabular}{ l  |  c   c   c   c} 
\toprule
Model  & Speed (FPS) & GPU Mem. (GB)  \\
\midrule
MinkLoc++    &  95  &   \first{2.0} \\
AdaFusion    &  84  &   \first{2.0} \\
UMF          &  90  &   \first{2.0} \\
LCPR         &  \first{110} &   2.1 \\
PRFusion (ours)     &  65  &   2.2 \\
PRFusion++ (ours)   &  40  &   2.3 \\
\bottomrule
\end{tabular}
\end{table}

\subsubsection{Run Time Speed and GPU Memory Usage}
Finally, we present the run time speed and GPU memory usage in \cref{tab:speed}. Compared with baselines, PRFusion and PRFusion++ are computationally costlier with the inclusion of the attention operation. However, these models still meet typical real-time deployment requirements (FPS $>30$). The higher computational cost can be mitigated with various solutions, such as knowledge distillation\cite{2021kd_survey}, quantization\cite{2018quantization}, and pruning\cite{2018pruning}, which are interesting future work for further improvements.

\section{Conclusion and Limitation}\label{sec:conclusion}
We have designed two multi-modal PR models, PRFusion and PRFusion++, which leverage manifold-based attention to facilitate more effective feature fusion at both global and local levels. Additionally, these models are equipped with neural Beltrami diffusion for more robust feature learning. The experimental results on large-scale benchmarks demonstrate the effectiveness of our design, as both models achieve SOTA performance.

However, due to separate feature extraction for 2D and 3D inputs and the additional computational overhead of attention computations, these models are computationally more expensive than several existing baselines. An interesting avenue for future work is to improve the pipeline for faster and lighter PR. To achieve this goal, we can explore various solutions, such as knowledge distillation, quantization, and pruning, which offer promising directions for further optimization.

\bibliographystyle{IEEEtran}
\bibliography{egbib}

\begin{thebibliography}{10}
\providecommand{\url}[1]{#1}
\csname url@samestyle\endcsname
\providecommand{\newblock}{\relax}
\providecommand{\bibinfo}[2]{#2}
\providecommand{\BIBentrySTDinterwordspacing}{\spaceskip=0pt\relax}
\providecommand{\BIBentryALTinterwordstretchfactor}{4}
\providecommand{\BIBentryALTinterwordspacing}{\spaceskip=\fontdimen2\font plus
\BIBentryALTinterwordstretchfactor\fontdimen3\font minus \fontdimen4\font\relax}
\providecommand{\BIBforeignlanguage}[2]{{%
\expandafter\ifx\csname l@#1\endcsname\relax
\typeout{** WARNING: IEEEtran.bst: No hyphenation pattern has been}%
\typeout{** loaded for the language `#1'. Using the pattern for}%
\typeout{** the default language instead.}%
\else
\language=\csname l@#1\endcsname
\fi
#2}}
\providecommand{\BIBdecl}{\relax}
\BIBdecl

\bibitem{wang2018navigation}
T.-H. Wang, H.-J. Huang, J.-T. Lin, C.-W. Hu, K.-H. Zeng, and M.~Sun, ``{Omnidirectional CNN for visual place recognition and navigation},'' in \emph{Proceedings of the IEEE International Conference on Robotics and Automation}, 2018, pp. 2341--2348.

\bibitem{doan2019autonomousdriving}
A.-D. Doan, Y.~Latif, T.-J. Chin, Y.~Liu, T.-T. Do, and I.~Reid, ``{Scalable place recognition under appearance change for autonomous driving},'' in \emph{Proceedings of the IEEE/CVF International Conference on Computer Vision}, 2019, pp. 9319--9328.

\bibitem{sarlin2022lamar}
P.-E. Sarlin, M.~Dusmanu, J.~L. Sch{\"o}nberger, P.~Speciale, L.~Gruber, V.~Larsson, O.~Miksik, and M.~Pollefeys, ``{LaMAR: Benchmarking localization and mapping for augmented reality},'' in \emph{Proceedings of the European Conference on Computer Vision}, 2022, pp. 686--704.

\bibitem{galvez2012bagofwords}
D.~G{\'a}lvez-L{\'o}pez and J.~D. Tardos, ``{Bags of binary words for fast place recognition in image sequences},'' \emph{IEEE Transactions on Robotics}, vol.~28, no.~5, pp. 1188--1197, 2012.

\bibitem{jegou2011vlad}
H.~J{\'e}gou, F.~Perronnin, M.~Douze, J.~S{\'a}nchez, P.~P{\'e}rez, and C.~Schmid, ``{Aggregating local image descriptors into compact codes},'' \emph{IEEE Transactions on Pattern Analysis and Machine Intelligence}, vol.~34, no.~9, pp. 1704--1716, 2011.

\bibitem{arandjelovic2016netvlad}
R.~Arandjelovic, P.~Gronat, A.~Torii, T.~Pajdla, and J.~Sivic, ``{NetVLAD: CNN architecture for weakly supervised place recognition},'' in \emph{Proceedings of the IEEE/CVF Conference on Computer Vision and Pattern Recognition}, 2016, pp. 5297--5307.

\bibitem{radenovic2018gempooling}
F.~Radenovi{\'c}, G.~Tolias, and O.~Chum, ``{Fine-tuning CNN image retrieval with no human annotation},'' \emph{IEEE Transactions on Pattern Analysis and Machine Intelligence}, vol.~41, no.~7, pp. 1655--1668, 2018.

\bibitem{berton2022cosplace}
G.~Berton, C.~Masone, and B.~Caputo, ``{Rethinking visual geo-localization for large-scale applications},'' in \emph{Proceedings of the IEEE/CVF Conference on Computer Vision and Pattern Recognition}, 2022, pp. 4878--4888.

\bibitem{ali2022convap}
A.~Alibey, B.~Chaibdraa, and P.~Giguere, ``{GSV-Cities: Toward appropriate supervised visual place recognition},'' \emph{Neurocomputing}, vol. 513, pp. 194--203, 2022.

\bibitem{ali2023mixvpr}
A.~Alibey and B.~Chaibdraa, ``{MixVPR: Feature mixing for visual place recognition},'' in \emph{Proceedings of the IEEE/CVF Winter Conference on Applications of Computer Vision}, 2023, pp. 2998--3007.

\bibitem{komorowski2021minkloc3d}
J.~Komorowski, ``{MinkLoc3d: Point cloud based large-scale place recognition},'' in \emph{Proceedings of the IEEE/CVF Winter Conference on Applications of Computer Vision}, 2021, pp. 1790--1799.

\bibitem{2021soenet}
Y.~Xia, Y.~Xu, S.~Li, R.~Wang, J.~Du, D.~Cremers, and U.~Stilla, ``{SOE-Net: A self-attention and orientation encoding network for point cloud based place recognition},'' in \emph{Proceedings of the IEEE/CVF Conference on computer vision and pattern recognition}, 2021, pp. 11\,348--11\,357.

\bibitem{minkloc3dv2}
K.~Jacek, ``{Improving point cloud based place recognition with ranking-based loss and large Batch Training},'' in \emph{Proceedings of the International Conference on Pattern Recognition}, 2022, pp. 3699--3705.

\bibitem{komorowski2021minkloc++}
J.~Komorowski, M.~Wysocza{\'n}ska, and T.~Trzcinski, ``{MinkLoc++: Lidar and monocular image fusion for place recognition},'' in \emph{Proceedings of the International Joint Conference on Neural Networks}, 2021, pp. 1--8.

\bibitem{lai2022adafusion}
H.~Lai, P.~Yin, and S.~Scherer, ``{AdaFusion: Visual-lidar fusion with adaptive weights for place recognition},'' \emph{IEEE Robotics and Automation Letters}, vol.~7, no.~4, pp. 12\,038--12\,045, 2022.

\bibitem{vaswani2017transformer}
A.~Vaswani, N.~Shazeer, N.~Parmar, J.~Uszkoreit, L.~Jones, A.~N. Gomez, {\L}.~Kaiser, and I.~Polosukhin, ``{Attention is all you need},'' in \emph{Advances in Neural Information Processing Systems}, 2017.

\bibitem{she2023image}
R.~She, Q.~Kang, S.~Wang, W.~P. Tay, Y.~L. Guan, D.~N. Navarro, and A.~Hartmannsgruber, ``Image patch-matching with graph-based learning in street scenes,'' \emph{IEEE Transactions on Image Processing}, vol.~32, pp. 3465--3480, 2023.

\bibitem{hausler2021patchnetvlad}
S.~Hausler, S.~Garg, M.~Xu, M.~Milford, and T.~Fischer, ``Patch-netvlad: Multi-scale fusion of locally-global descriptors for place recognition,'' in \emph{Proceedings of the IEEE/CVF Conference on Computer Vision and Pattern Recognition}, 2021, pp. 14\,141--14\,152.

\bibitem{cai2022patchnetvlad+}
Y.~Cai, J.~Zhao, J.~Cui, F.~Zhang, T.~Feng, and C.~Ye, ``Patch-netvlad+: Learned patch descriptor and weighted matching strategy for place recognition,'' in \emph{Proceedings of the IEEE International Conference on Multisensor Fusion and Integration for Intelligent Systems}, 2022, pp. 1--8.

\bibitem{wang2022transvpr}
R.~Wang, Y.~Shen, W.~Zuo, S.~Zhou, and N.~Zheng, ``Transvpr: Transformer-based place recognition with multi-level attention aggregation,'' in \emph{Proceedings of the IEEE/CVF Conference on Computer Vision and Pattern Recognition}, 2022, pp. 13\,648--13\,657.

\bibitem{zhu2023r2former}
S.~Zhu, L.~Yang, C.~Chen, M.~Shah, X.~Shen, and H.~Wang, ``R2former: Unified retrieval and reranking transformer for place recognition,'' in \emph{Proceedings of the IEEE/CVF Conference on Computer Vision and Pattern Recognition}, 2023, pp. 19\,370--19\,380.

\bibitem{Wang2023HypLiLoc}
S.~Wang, Q.~Kang, R.~She, W.~Wang, K.~Zhao, Y.~Song, and W.~P. Tay, ``Hypliloc: Towards effective lidar pose regression with hyperbolic fusion,'' in \emph{Proceedings of the IEEE/CVF Conference on Computer Vision and Pattern Recognition}, 2023, pp. 5176--5185.

\bibitem{li2023sgloc}
W.~Li, S.~Yu, C.~Wang, G.~Hu, S.~Shen, and C.~Wen, ``Sgloc: Scene geometry encoding for outdoor lidar localization,'' in \emph{Proceedings of the IEEE/CVF Conference on Computer Vision and Pattern Recognition}, 2023, pp. 9286--9295.

\bibitem{uy2018pointnetvlad}
M.~A. Uy and G.~H. Lee, ``{PointNetVLAD: Deep point cloud based retrieval for large-scale place recognition},'' in \emph{Proceedings of the IEEE/CVF Conference on Computer Vision and Pattern Recognition}, 2018, pp. 4470--4479.

\bibitem{qi2017pointnet}
C.~R. Qi, H.~Su, K.~Mo, and L.~J. Guibas, ``{Pointnet: Deep learning on point sets for 3d classification and segmentation},'' in \emph{Proceedings of the IEEE/CVF Conference on Computer Vision and Pattern Recognition}, 2017, pp. 652--660.

\bibitem{qi2017pointnet++}
C.~R. Qi, L.~Yi, H.~Su, and L.~J. Guibas, ``{Pointnet++: Deep hierarchical feature learning on point sets in a metric space},'' in \emph{Advances in Neural Information Processing Systems}, 2017, pp. 1--10.

\bibitem{zhang2019pcan}
W.~Zhang and C.~Xiao, ``{PCAN: 3D attention map learning using contextual information for point cloud based retrieval},'' in \emph{Proceedings of the IEEE/CVF Conference on Computer Vision and Pattern Recognition}, 2019, pp. 12\,436--12\,445.

\bibitem{du2020dh3d}
J.~Du, R.~Wang, and D.~Cremers, ``{DH3d: Deep hierarchical 3d descriptors for robust large-scale 6dof relocalization},'' in \emph{Proceedings of the European Conference on Computer Vision}, 2020, pp. 744--762.

\bibitem{2019lpdnet}
Z.~Liu, S.~Zhou, C.~Suo, P.~Yin, W.~Chen, H.~Wang, H.~Li, and Y.-H. Liu, ``{LPD-net: 3d point cloud learning for large-scale place recognition and environment analysis},'' in \emph{Proceedings of the IEEE/CVF International Conference on Computer Vision}, 2019, pp. 2831--2840.

\bibitem{2022epcnet}
L.~Hui, M.~Cheng, J.~Xie, J.~Yang, and M.-M. Cheng, ``{Efficient 3D point cloud feature learning for large-scale place recognition},'' \emph{IEEE Transactions on Image Processing}, vol.~31, pp. 1258--1270, 2022.

\bibitem{chen2023ptcnet}
L.~Chen, H.~Wang, H.~Kong, W.~Yang, and M.~Ren, ``{PTC-Net: Point-wise transformer with sparse convolution network for place recognition},'' \emph{IEEE Robotics and Automation Letters}, vol.~8, no.~6, pp. 3414--3421, 2023.

\bibitem{yin2021psematch}
P.~Yin, L.~Xu, Z.~Feng, A.~Egorov, and B.~Li, ``Pse-match: A viewpoint-free place recognition method with parallel semantic embedding,'' \emph{IEEE Transactions on Intelligent Transportation Systems}, vol.~23, no.~8, pp. 11\,249--11\,260, 2021.

\bibitem{zywanowski2021minkloc3dsi}
K.~{\.Z}ywanowski, A.~Banaszczyk, M.~R. Nowicki, and J.~Komorowski, ``Minkloc3d-si: 3d lidar place recognition with sparse convolutions, spherical coordinates, and intensity,'' \emph{IEEE Robotics and Automation Letters}, vol.~7, no.~2, pp. 1079--1086, 2021.

\bibitem{ma2022overlaptransformer}
J.~Ma, J.~Zhang, J.~Xu, R.~Ai, W.~Gu, and X.~Chen, ``Overlaptransformer: An efficient and yaw-angle-invariant transformer network for lidar-based place recognition,'' \emph{IEEE Robotics and Automation Letters}, vol.~7, no.~3, pp. 6958--6965, 2022.

\bibitem{luo2023bevplace}
L.~Luo, S.~Zheng, Y.~Li, Y.~Fan, B.~Yu, S.-Y. Cao, J.~Li, and H.-L. Shen, ``Bevplace: Learning lidar-based place recognition using bird's eye view images,'' in \emph{Proceedings of the IEEE/CVF International Conference on Computer Vision}, 2023, pp. 8700--8709.

\bibitem{lu2020picnet}
Y.~Lu, F.~Yang, F.~Chen, and D.~Xie, ``{PIC-Net: Point cloud and image collaboration network for large-scale place recognition},'' \emph{arXiv preprint arXiv:2008.00658}, 2020.

\bibitem{oertel2020cues}
A.~Oertel, T.~Cieslewski, and D.~Scaramuzza, ``{Augmenting visual place recognition with structural cues},'' \emph{IEEE Robotics and Automation Letters}, vol.~5, no.~4, pp. 5534--5541, 2020.

\bibitem{zhou2023lcpr}
Z.~Zhou, J.~Xu, G.~Xiong, and J.~Ma, ``Lcpr: a multi-scale attention-based lidar-camera fusion network for place recognition,'' \emph{IEEE Robotics and Automation Letters}, vol.~9, no.~2, pp. 1342--1349, 2024.

\bibitem{garcia2024umf}
A.~Garc{\'\i}a-Hern{\'a}ndez, R.~Giubilato, K.~H. Strobl, J.~Civera, and R.~Triebel, ``Unifying local and global multimodal features for place recognition in aliased and low-texture environments,'' \emph{arXiv preprint arXiv:2403.13395}, 2024.

\bibitem{chen2018ode}
R.~T. Chen, Y.~Rubanova, J.~Bettencourt, and D.~K. Duvenaud, ``{Neural ordinary differential equations},'' in \emph{Advances in Neural Information Processing Systems}, 2018, pp. 1--13.

\bibitem{yan2019tisode}
H.~Yan, J.~Du, V.~Y. Tan, and J.~Feng, ``{On robustness of neural ordinary differential equations},'' \emph{arXiv preprint arXiv:1910.05513}, 2019.

\bibitem{kang2021sodef}
Q.~Kang, Y.~Song, Q.~Ding, and W.~P. Tay, ``{Stable neural ODE with Lyapunov-stable equilibrium points for defending against adversarial attacks},'' in \emph{Advances in Neural Information Processing Systems}, 2021, pp. 14\,925--14\,937.

\bibitem{chamberlain2021grand}
B.~Chamberlain, J.~Rowbottom, M.~I. Gorinova, M.~Bronstein, S.~Webb, and E.~Rossi, ``{Grand: Graph neural diffusion},'' in \emph{Proceedings of the International Conference on Machine Learning}, 2021, pp. 1407--1418.

\bibitem{chamberlain2021blend}
B.~Chamberlain, J.~Rowbottom, D.~Eynard, F.~Di~Giovanni, X.~Dong, and M.~Bronstein, ``{Beltrami flow and neural diffusion on graphs},'' in \emph{Advances in Neural Information Processing Systems}, 2021, pp. 1594--1609.

\bibitem{zhao2023adversarial}
K.~Zhao, Q.~Kang, Y.~Song, R.~She, S.~Wang, and W.~P. Tay, ``{Adversarial robustness in graph neural networks: A Hamiltonian approach},'' \emph{arXiv preprint arXiv:2310.06396}, 2023.

\bibitem{she2023robustmat}
R.~She, Q.~Kang, S.~Wang, Y.-R. Yang, K.~Zhao, Y.~Song, and W.~P. Tay, ``{RobustMat: Neural diffusion for street landmark patch matching under challenging environments},'' \emph{IEEE Transactions on Image Processing}, vol.~32, pp. 5550--5563, 2023.

\bibitem{wang2023robustloc}
S.~Wang, Q.~Kang, R.~She, W.~P. Tay, A.~Hartmannsgruber, and D.~N. Navarro, ``{RobustLoc: Robust camera pose regression in challenging driving environments},'' in \emph{Proceedings of the AAAI Conference on Artificial Intelligence}, 2023, pp. 6209--6216.

\bibitem{song2022robustness}
Y.~Song, Q.~Kang, S.~Wang, K.~Zhao, and W.~P. Tay, ``{On the robustness of graph neural diffusion to topology perturbations},'' in \emph{Advances in Neural Information Processing Systems}, 2022, pp. 6384--6396.

\bibitem{beltrami1}
V.~S. Matveev, ``Geometric explanation of the beltrami theorem,'' \emph{International Journal of Geometric Methods in Modern Physics}, vol.~3, no.~3, p. 623, 2006.

\bibitem{beltrami2}
J.~Etnyre and R.~Ghrist, ``Contact topology and hydrodynamics: I. beltrami fields and the seifert conjecture,'' \emph{Nonlinearity}, vol.~13, no.~2, p. 441, 2000.

\bibitem{beltrami3}
U.~Boscain and C.~Laurent, ``The laplace-beltrami operator in almost-riemannian geometry,'' \emph{Annales de l'institut Fourier}, vol.~63, no.~5, pp. 1739--1770, 2013.

\bibitem{beltrami4}
Z.~Yoshida and S.~Mahajan, ``Simultaneous beltrami conditions in coupled vortex dynamics,'' \emph{Journal of Mathematical Physics}, vol.~40, no.~10, pp. 5080--5091, 1999.

\bibitem{dosovitskiy2020vit}
A.~Dosovitskiy, L.~Beyer, A.~Kolesnikov, D.~Weissenborn, X.~Zhai, T.~Unterthiner, M.~Dehghani, M.~Minderer, G.~Heigold, S.~Gelly \emph{et~al.}, ``{An image is worth 16x16 words: Transformers for image recognition at scale},'' \emph{arXiv preprint arXiv:2010.11929}, 2020.

\bibitem{lai2023spherical}
X.~Lai, Y.~Chen, F.~Lu, J.~Liu, and J.~Jia, ``{Spherical transformer for LiDAR-based 3D recognition},'' \emph{arXiv preprint arXiv:2303.12766}, 2023.

\bibitem{liu2021swin}
Z.~Liu, Y.~Lin, Y.~Cao, H.~Hu, Y.~Wei, Z.~Zhang, S.~Lin, and B.~Guo, ``{Swin transformer: Hierarchical vision transformer using shifted windows},'' in \emph{Proceedings of the IEEE International Conference on Computer Vision}, 2021, pp. 10\,012--10\,022.

\bibitem{hoffer2015tripletloss}
E.~Hoffer and N.~Ailon, ``{Deep metric learning using triplet network},'' in \emph{International Workshop on Similarity-Based Pattern Recognition}, 2015, pp. 84--92.

\bibitem{kingma2014adam}
D.~P. Kingma and J.~Ba, ``{Adam: A method for stochastic optimization},'' \emph{arXiv preprint arXiv:1412.6980}, 2014.

\bibitem{oxford}
W.~Maddern, G.~Pascoe, C.~Linegar, and P.~Newman, ``{1 year, 1000 km: The Oxford RobotCar dataset},'' \emph{The International Journal of Robotics Research}, vol.~36, no.~1, pp. 3--15, 2017.

\bibitem{geiger2013kitti}
A.~Geiger, P.~Lenz, C.~Stiller, and R.~Urtasun, ``{Vision meets robotics: The KITTI dataset},'' \emph{The International Journal of Robotics Research}, vol.~32, no.~11, pp. 1231--1237, 2013.

\bibitem{burnett2022boreas}
K.~Burnett, D.~J. Yoon, Y.~Wu, A.~Z. Li, H.~Zhang, S.~Lu, J.~Qian, W.-K. Tseng, A.~Lambert, K.~Y. Leung \emph{et~al.}, ``{Boreas: A multi-season autonomous driving dataset},'' \emph{arXiv preprint arXiv:2203.10168}, 2022.

\bibitem{pan2021coral}
Y.~Pan, X.~Xu, W.~Li, Y.~Cui, Y.~Wang, and R.~Xiong, ``{CORAL: Colored structural representation for bi-modal place recognition},'' in \emph{Proceedings of the IEEE/RSJ International Conference on Intelligent Robots and Systems}, 2021, pp. 2084--2091.

\bibitem{zhang2004extrinsic}
Q.~Zhang and R.~Pless, ``Extrinsic calibration of a camera and laser range finder (improves camera calibration),'' in \emph{Proceedings of the IEEE/RSJ International Conference on Intelligent Robots and Systems}, vol.~3, 2004, pp. 2301--2306.

\bibitem{zhou2018automatic}
L.~Zhou, Z.~Li, and M.~Kaess, ``Automatic extrinsic calibration of a camera and a 3d lidar using line and plane correspondences,'' in \emph{Proceedings of the IEEE/RSJ International Conference on Intelligent Robots and Systems}, 2018, pp. 5562--5569.

\bibitem{schneider2017regnet}
N.~Schneider, F.~Piewak, C.~Stiller, and U.~Franke, ``Regnet: Multimodal sensor registration using deep neural networks,'' in \emph{Proceedings of the IEEE Intelligent Vehicles Symposium}, 2017, pp. 1803--1810.

\bibitem{iyer2018calibnet}
G.~Iyer, R.~K. Ram, J.~K. Murthy, and K.~M. Krishna, ``Calibnet: Geometrically supervised extrinsic calibration using 3d spatial transformer networks,'' in \emph{Proceedings of the IEEE/RSJ International Conference on Intelligent Robots and Systems}, 2018, pp. 1110--1117.

\bibitem{lv2021lccnet}
X.~Lv, B.~Wang, Z.~Dou, D.~Ye, and S.~Wang, ``Lccnet: Lidar and camera self-calibration using cost volume network,'' in \emph{Proceedings of the IEEE/CVF Conference on Computer Vision and Pattern Recognition}, 2021, pp. 2894--2901.

\bibitem{2021kd_survey}
J.~Gou, B.~Yu, S.~J. Maybank, and D.~Tao, ``Knowledge distillation: A survey,'' \emph{International Journal of Computer Vision}, vol. 129, pp. 1789--1819, 2021.

\bibitem{2018quantization}
A.~Polino, R.~Pascanu, and D.~Alistarh, ``Model compression via distillation and quantization,'' \emph{arXiv preprint arXiv:1802.05668}, 2018.

\bibitem{2018pruning}
Z.~Liu, M.~Sun, T.~Zhou, G.~Huang, and T.~Darrell, ``Rethinking the value of network pruning,'' \emph{arXiv preprint arXiv:1810.05270}, 2018.

\end{thebibliography}

\end{document}